%% file: main.tex
\documentclass[conference]{IEEEtran}
\IEEEoverridecommandlockouts

\usepackage{amsmath,amsfonts}
\usepackage{algorithm}
\usepackage{algorithmicx}
\usepackage{algpseudocode}
\usepackage{array}
\usepackage[caption=false,font=normalsize,labelfont=sf,textfont=sf]{subfig}
\usepackage{textcomp}
\usepackage{multirow}
\usepackage{stfloats}
\usepackage{url}
\usepackage{bbm}
\usepackage{verbatim}
\usepackage{graphicx}
\usepackage{threeparttable}
\def\BibTeX{{\rm B\kern-.05em{\sc i\kern-.025em b}\kern-.08em
    T\kern-.1667em\lower.7ex\hbox{E}\kern-.125emX}}
\usepackage{balance}
\begin{document}
\title{Target-agnostic Source-free Domain Adaptation for Regression Tasks}

\author{
\IEEEauthorblockN{Tianlang He${}^1$, Zhiqiu Xia${}^2$, Jierun Chen${}^1$, Haoliang Li${}^3$, S.-H. Gary Chan${}^1$ }
\IEEEauthorblockA{${}^{1,2}$Department of Computer Science and Engineering,
The Hong Kong University of Science and Technology \\
% Hong Kong, China\\
%\and
%\IEEEauthorblockN{}
\IEEEauthorblockA{${}^3$Department of Electronic Engineering, 
City University of Hong Kong\\
% Hong Kong, China\\
Email: ${}^1$\{theaf, jcheneh, gchan\}@cse.ust.hk, ${}^2$zxiaae@connect.ust.hk, ${}^3$haoliang.li@cityu.edu.hk}
}
}
%\author{Anonymous Authors

%\title{SODAR: Source-free Deep Regressor Adaptation for Resource-constrained Devices}
%\author{Tianlang He, Zhiqiu Xia, Jierun Chen, Haoliang Li, S.-H.~Gary Chan
%\thanks{Manuscript created October, 2020; This work was developed by the IEEE Publication Technology Department. This work is distributed under the \LaTeX \ Project Public License (LPPL) ( http://www.latex-project.org/ ) version 1.3. A copy of the LPPL, version 1.3, is included in the base \LaTeX \ documentation of all distributions of \LaTeX \ released 2003/12/01 or later. The opinions expressed here are entirely that of the author. No warranty is expressed or implied. User assumes all risk. (Corresponding author: Tianlang He)}

%\thanks{Tianlang He, Zhiqiu Xia, Jierun Chen, and S.-H. Gary Chan are with the Computer Science and Engineering Department, The Hong Kong University of Science and Technology, Clear Water Bay, N.T., Hong Kong S.A.R. (e-mail: theaf@cse.ust.hk)}
%\thanks{Haoliang Li is with the Department of Electronic Engineering, City University of Hong Kong, Tat Chee Avenue, Kowloon, Hong Kong S.A.R.}
%}

%\markboth{Journal of \LaTeX\ Class Files,~Vol.~18, No.~9, September~2020}%
%{How to Use the IEEEtran \LaTeX \ Templates}

\newcommand*{\sysname}{TASFAR}

\maketitle

\input{section/abstract}


\input{section/introduction.tex}

\input{section/relate.tex}

\input{section/method.tex}

\input{section/exp.tex}

\input{section/conclude.tex}
\input{section/discuss.tex}

\bibliographystyle{IEEEtran}
\bibliography{citation}

% Upload this in the end
\iffalse
\begin{IEEEbiographynophoto}{Jane Doe}
Biography text here without a photo.
\end{IEEEbiographynophoto}

\begin{IEEEbiography}[{\includegraphics[width=1in,height=1.25in,clip,keepaspectratio]{fig1.png}}]{IEEE Publications Technology Team}
In this paragraph you can place your educational, professional background and research and other interests.\end{IEEEbiography}

\fi

\end{document}

%% file: section/abstract.tex
\begin{abstract}

% Unsupervised Domain Adaptation (UDA) is a learning framework to transfer knowledge from source domains with a large number of annotated training examples to target domains with unlabeled data only
% Traditional UDA requires labeled data from the source domain and unlabeled data from the target domain to improve performance (prediction accuracy) in the target domain

% Problem 
% 1:
% We consider unsupervised domain adaptation~(UDA), where unlabeled target data are used as training data to address the domain gap between the source and target. Traditional approaches require training
% datasets of the source as well for such adaptation.
%Traditional unsupervised domain adaptation~(UDA) seeks to bridge the domain gap between the target and source using unlabeled target data and labeled source data.
Unsupervised domain adaptation~(UDA) seeks to bridge the domain gap between the target and the source using unlabeled target data.  
%Traditional approaches often require labeled source data at the target (source-based), raising concerns on data confidentiality and storage overhead.
% 2: source-free UDA requires a source model, pre-trained by source datasets, and unlabeled target data to achieve the goal of traditional UDA
%To remove the source data, recent source-free UDA approaches  
% {\LARGE A: PLEASE MAKE IT PARALLEL WITH B BELOW: aligning the domain gap in input or feature space}, 
%extract domain-invariant features from the domain gap estimated by a classifier, which unfortunately 
% degrades performance and 
%limits to 
%
Source-free UDA removes the requirement for labeled source data at the target to preserve data privacy and storage.  
However, work on source-free UDA assumes knowledge of domain gap distribution, 
and hence is limited to either target-aware or classification task.
% However, prior work on source-free UDA is based on domain gap estimation, and hence is limited to either classification tasks or specific target domains.  
%
To overcome it,
% Approach
% 4
we propose \sysname{}, a novel {\bf t}arget-{\bf a}gnostic {\bf s}ource-{\bf f}ree domain {\bf a}daptation approach for {\bf r}egression tasks.
% 5
Using prediction confidence,
\sysname{} estimates a label density map as the target label distribution, which is then used to calibrate the source model on the target domain.
%that  operates on the input or feature space of the source model,
% \sysname{} {\LARGE B: PLEASE CONTRAST IT WITH A ABOVE directly aligns the gap between source model predictions and target label distribution},
%directly optimizes the predictions of target data by the source model 
%achieving adaptation without assumptions about target-specific information or classification tasks.
% Result
% 6
We have conducted extensive experiments on four regression tasks with various domain gaps, namely,
pedestrian dead reckoning for different users, 
image-based people counting in different scenes,
housing-price prediction at different districts,
and taxi-trip duration prediction from different departure points.
% We conduct extensive experiments on the regression tasks of pedestrian dead reckoning~(based on inertial measurement unit data) and image-based people counting with diverse target users and crowd scenes, and two prediction tasks that validates \sysname{}'s generality. 
\sysname{} is shown to
%{\LARGE PLEASE CHECK CORRECTNESS
substantially outperform the state-of-the-art source-free UDA approaches by
averagely reducing 22\% errors for the four tasks and
% significant reduction in localization errors (about 14\%) or MSE (about 24\%), and
%based on pre-defined domain gap, 
achieve notably comparable accuracy as source-based UDA without using source data.
%}
%, without prior knowledge of source data and target domain.
% {\LARGE HOW ABOUT TARGET DOMAIN STUDY?} 

\end{abstract}

\iffalse

 Traditional unsupervised domain adaptation~(UDA) requires training
datasets of the source, alongside with unlabeled target data, to address the covariate shift between the source
and target domains. To better protect
data privacy, source-free domain adaptation~(SFDA) uses source
models instead of source datasets to adapt to target domains.
%and unlabeled target data to improve target
%accuracy, which avoids data sharing and fits more practical
%application scenarios.
%
However, existing SFDA design applies
%on the characteristics of
to classification tasks only and can be hardly extended to
regression tasks.  We propose, for the first time,
\sysname{}, a simple implementable \emph{\textbf{so}}urce-free
\emph{\textbf{d}}omain \emph{\textbf{a}}daptation approach for general
\emph{\textbf{r}}egression tasks using Baysian neural networks.
Based on the prediction 
uncertainty of the target data,
\sysname{}
constructs a target label space which is used to
adapt the source
model to the target one by means of a domain label loss function.
%To the best of our knowledge, \sysname{} is the first
%general SFDA approach to tackle the covariate shift in regression
%tasks.
%\sysname{} is general to regression tasks and easy to
%implement.
Through extensive experiments on regression tasks for location sensing and
people counting, we show that \sysname{} performance is comparable with traditional UDAs, without the need for source dataset.

\fi

%% file: section/introduction.tex
\section{Introduction}\label{sec:intro}

% Context: deep regression tasks, adaptation 
\IEEEPARstart{D}{eep} learning has been demonstrated with promising results in many tasks, 
such as location sensing~\cite{wu2019efficient}~\cite{proxihe}, 
 people counting~\cite{wang2020nwpu}~\cite{sindagi2018survey}, activity recognition~\cite{ramanujam2021human}~\cite{wilson2022domain}, etc.
%To protect data privacy, these tasks are usually conducted on edge devices of users. %, say, smartphones. 
Despite that, the performance of deep models often degrades significantly 
when the target data shift from  
the input distribution of the training dataset (i.e.~the source domain). %, the so-called covariate shifts.
To tackle this,
unsupervised domain adaptation~(UDA) has been proposed to learn a target model, i.e.~aligning the model features extracted from both source and target domains using unlabelled target data.
%to
% extract similar deep features from the target and source,
%restoring the accuracy in targets to that of the source. 

% Source-free
In traditional UDA, the source dataset is made available to the target for adaptation.  As the database may be large (several gigabytes or more), this is not storage-efficient and not applicable for
storage-constrained devices. % may cite some devices in here
Though some works~\cite{kurmi2021domain}~\cite{ICML2022Measurement} have tried to reduce such storage by 
compression, % with deep generative models~\cite{kurmi2021domain, li2020model} or source feature statistics~\cite{ICML2022Measurement, fan2022unsupervised}.
they still consume substantial storage with trade-off on the adaptation quality. To remove dataset storage,
source-free UDA has been recently proposed, which is to adapt the source model pre-trained by source data directly with 
unlabeled target data.
% We do not consider the issues after the first-time adaptation
In this work, we consider one-shot source-free UDA;
readers interested in consecutive adaptations may refer to studies on continual learning~\cite{hadsell2020embracing}~\cite{de2021continual} and references therein. 

% Basic idea of source-free approach and existing arts 
In source-free UDA, the absence of source data  causes complication
in measuring the domain gap (the discrepancy between the input distribution of the source and target).
Existing approaches bridge the gap based on either input data or features. 
The approaches based on input data require prior knowledge of domain gap and 
simulate such gap by means of data augmentation to extract invariant features~\cite{yu2022source}~\cite{xiong2021source}~\cite{karim2023c}.
On the other hand, the feature-based approach is generally applied to classification tasks by
measuring and minimizing domain gaps in feature space~\cite{yang2021generalized}~\cite{kundu2020universal}~\cite{yang2021exploiting}~\cite{zhang2022divide}.
%leveraging classification properties
%such as information entropy of classification scores~(or prediction confidence)~\cite{yang2021generalized,kundu2020universal} and
%category-based feature compactness~\cite{yang2021exploiting,zhang2022divide}.
% Problem

While impressive, 
previous approaches assume the knowledge of domain gap.
% What does the "target-agnostic" mean?
As target domains are often agnostic when designing adaptation algorithms, 
% such 
domain gaps can be unknown, heterogeneous and complex to simulate~\cite{wang2023leto}. 
% e.g., 
For example, a source model may be deployed in different target scenarios in terms of user behaviors, device heterogeneity and operating environments. 
% agnostic targets usually make the domain gaps heterogeneous and complex. 
%A source model may also be deployed in different target scenarios of different users and environments.
%Data from different target scenarios may differ from the source data in heterogeneous aspects, 
%rendered by various and unexpected causes such as
%heterogeneous user behaviors, environments, devices, and so on.
% And the reason why we cannot use input space to adapt
%Also,
%some domain gaps can be very complicated to simulate; for example,
%the environmental impact on wireless signals~\cite{wang2023leto} 
%is still a tricky problem 
%after decades of studies.
%Thus,
%the agnostic target rigidly demands an adaptation approach
%independent of specific domain gaps.
Furthermore, many machine learning tasks are regression in nature.  
% How does regression cause trouble?
In contrast to classification tasks where target data of the same label can be correlated in the feature space to shed light on domain gap~\cite{yang2021exploiting}, 
the vast continuous label space of regression tasks without overlapping labels
poses great difficulty for the deep regressor to adapt and converge.
%because the target data barely share identical labels,
%making the extracted features decoupled and independent. 
% invalidates measuring the domain gap in feature space, which
%as compared to the classification model.
% poses great challenges to adaptation for regressors.
%Since the label space of classification is finite,
%target data sharing the same label can be correlated in the feature space;
%such feature correlation can shed light in measuring the domain gap~(such as feature compactness~\cite{yang2021exploiting}).
%Due to the continuous label space of regression tasks, however,
%target data barely share identical labels,
%making the extracted features decoupled and independent. 
%This frustrates the previous approaches in measuring domain gaps
%and calls for a new approach.
%Therefore, while impressive, the previous works 
%can hardly be extended to regression tasks 
%whose target domains are unknown beforehand.% ~(or agnostic).
%the target domains are usually unknown beforehand,
% when the adaptation module is built and delivered,
% which impedes previous approaches extended to regression tasks with such agnostic target domains.

% Our observation: alignment using label
%Although the data and features are hard to use for adaptation, 
We consider, for the first time, target-agnostic source-free UDA for regression tasks.
The key observation is that  
% the target labels, similar to its input data, are {\LARGE CHECK GRAMMAR produced from the target domain and hence may be correlated to the target scenarios}. 
the target label, like the input data that all conform to the target domain, also originates from the same target scenario.
As an example, if a target user's stride length mostly falls into a certain range~(say, 0.5 to 0.8m), 
his/her next stride length is likely within the range as well.
% {\LARGE SOMETHING WRONG}
Therefore,
in contrast to the previous approaches that measure and bridge domain gaps in input data or feature space of the source model,
% we employ the label distribution of the target domain to optimize the source model predictions on target data.  
we directly estimate the label distribution of the target scenario and use it to calibrate source models. 
By considering only the label distribution, we can achieve target-agnostic adaptation for regression tasks that is orthogonal to target domain.

\begin{figure}[!t]
\centering
\includegraphics[width=3.2in]{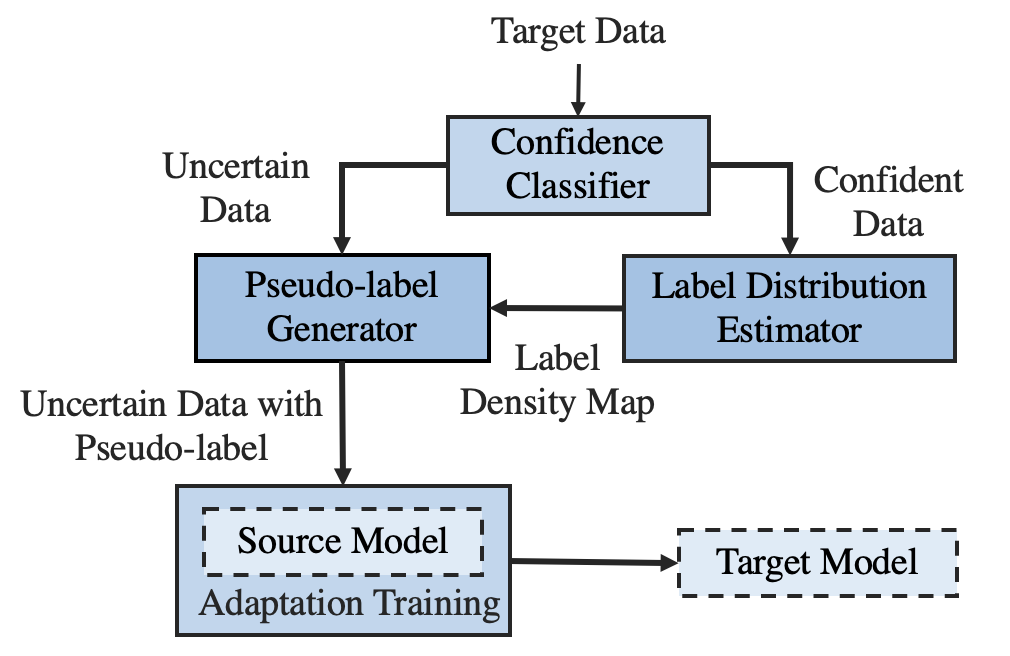}
\caption{System diagram of \sysname{}. 
        First, \sysname{} uses a confidence classifier to classify the target data into confident and uncertain data.
        % Target data are classified into confident data and uncertain data using a confidence classifier. 
        The confident data are used by a label distribution estimator to generate a label density map.
        The uncertain data are pseudo-labeled by a proposed pseudo-label generator based on the label density map. 
        Then,
        \sysname{} uses the pseudo-labeled uncertain data to train the source model to be the target model. 
        % {\LARGE HOW COME SOURCE MODEL AND TARGET MODEL ARE TRAIN TOGETHER?  SHOULDN'T THE OUTPUT OF THE SOURCE MODEL AFTER TRAINING BE TARGET MODEL?}
        }
\label{fig: sys_diagram}
\end{figure}

% First-level detail of the approach
We propose \emph{\textbf{\sysname{}}}, 
a novel \emph{\textbf{t}}arget-\emph{\textbf{a}}gnostic \emph{\textbf{s}}ource-\emph{\textbf{f}}ree domain \emph{\textbf{a}}daptation approach for  \emph{\textbf{r}}egression tasks. We 
%a general and implementable \emph{\textbf{so}}urce-free \emph{\textbf{d}}eep regressor \emph{\textbf{a}}daptation approach for \emph{\textbf{r}}esource-constrained devices. 
show its overall system diagram in Figure~\ref{fig: sys_diagram}.
\sysname{} first classifies the target data into confident data and uncertain data based on a confidence classifier depending on prediction confidence~\cite{gal2016uncertainty}~\cite{gawlikowski2021survey}.
Based on the confident data,
\sysname{} uses a label distribution estimator to generate a label density map.
% the confident data are used to generate a label density map by a label distribution estimator.
Then, a pseudo-label generator leverages the label density map 
to pseudo-label the uncertain data.
Finally, \sysname{} uses the pseudo-labeled uncertain data to fine-tune the source model by supervised learning,
after which the target model is delivered.

To the best of our knowledge, 
\sysname{} is the first target-agnostic source-free regressor adaptation approach based on label distribution.
Our contributions are the following:

\begin{itemize}
\item{\emph{Label distribution estimator using prediction confidence:} 
    % Since the labels come from the same target scenario, 
    % their distribution can be used for model adaptation. 
    We estimate the label distribution of confident data to pseudo-label uncertain data. 
    However, the target labels are unavailable in the setting of UDA. 
    We thus propose a label distribution estimator to overcome it.
    To be specific, 
    the proposed estimator utilizes the prediction confidence of the source model 
    to estimate the target label distribution, which is represented as a label density map.
    %    estimating the label distribution 
   % using the source model prediction.
    %, with the prediction confidence, on the confident data.
   % Specifically, our proposed  first 
  %{\LARGE THESE SENTENCES ARE VERBOSE AND CLUMSY  It   leverages the correlation between prediction error and confidence using the source model 
  %  to estimate the label distribution of the confident data. 
  %  It then summarizes the estimated instance-label distribution 
  %  to represent the label distribution of the target, which is termed the label density map. }
  %  }
\item{\emph{Pseudo-label generator based on label density map: } %that reduces the discrepancy between source model predictions and label density map: }    
    %Since all target labels are generated from the same target scenario, 
    % the labels of uncertain data should conform to the target label distribution. 
    Generated from the same target scenarios,
    the label distribution of confident data can be the prior knowledge of the labels of uncertain data. 
    Therefore, we propose a pseudo-label generator that utilizes the label density map to pseudo-label the uncertain data. 
    % calibrate the predictions of the source model on uncertain data.
    Specifically, 
    the pseudo-label generator pseudo-labels uncertain data by considering the joint distribution of label density map and source model prediction. 
    % the label density map and the source model prediction to generate pseudo-labels for uncertain data, 
    % such that the resulting pseudo-labels are closer to the label distribution of the target scenario.
    To avoid generating low-quality pseudo-labels that cause accuracy degradation, 
    the generator also weighs the pseudo-labels by evaluating their credibility based on the map densities. 
    %Produced from the same target scenario, the labels of the target 
    %tend to appear in the 
    % {\LARGE DENSELY APPEARED?  regions that densely appeared}.
   % {\LARGE VERY DIFFICULT TO UNDERSTND MAINLY DUE TO WORDY ENGLISH Thus, 
    % we regard it as a discrepancy 
    % when a source model prediction on the uncertain data appears in a region that is rarely observed on the label density map.
    % Based on {\LARGE WHAT INTUITION? this intuition}, 
    % we propose a pseudo-label generator that 
    % assigns pseudo-labels to the uncertain data so that the pseudo-labels are closer to the dense regions on the label density map than the source model predictions. 
    % 
    }
}   

\end{itemize}

We have conducted extensive experiments to validate \sysname{} on four regression tasks -- location sensing~(pedestrian dead reckoning)~\cite{yan2019ronin}, image-based people counting~\cite{zhang2016single}, and two prediction tasks~\cite{calihouseprice, nyc-taxi-trip-duration} --
and compared it with the existing source-free UDA with pre-defined domain gap and traditional source-based UDA approaches (expectedly the best performance due to the availability of source dataset). 
Our experimental results show that, as compared with the state-of-the-art source-free UDA approaches,
\sysname{} achieves on average a substantial $14\%$ and $24\%$ reduction in localization error on different users and counting MSE on various crowd scenes, respectively, and 22\% and 28\% reduction of prediction errors on the two prediction tasks.  Without access to source datasets, \sysname{} impressively achieves similar accuracy as the source-based UDA approaches.

The remainder of this paper is organized as follows.
We review related works in Section~\ref{sec: relate}
and present \sysname{} in Section~\ref{sec: method} in terms of its confidence classifier, label distribution estimator, and pseudo-label generator. 
%where 
%we overview the approach of \sysname{} in Section~\ref{subsec: overview} and discuss confidence %classification in Section~\ref{subsec: confidence_class}, and
%delineate label distribution estimator and target label loss function in Section~\ref{subsec: %estimator} and Section~\ref{subsec: loss}, respectively.
We discuss in Section~\ref{sec: exp}  illustrative experimental results, 
followed by conclusion in Section~\ref{sec: conclude}.
Finally, we discuss future works in Section~\ref{sec: discuss}.

%% file: section/relate.tex
\section{Related Work}\label{sec: relate}

%In this section, we review the literature in terms of traditional unsupervised domain adaptation~(UDA), UDA with source data %compression, and source-free UDA.

% First cut: traditional UDA
UDA for deep models has been extensively studied~\cite{zhao2020review}~\cite{DAsurvey}~\cite{farahani2021brief}.
These works align the source and target domains either by input data or deep model features.
Some pioneering works on data alignment~\cite{huang2006correcting}~\cite{phillips2009sample} reduce the domain gap by importance sampling on source data to simulate target data distribution.
Recent data alignment approaches~\cite{benaim2017one}~\cite{fu2019geometry}~\cite{hoffman2018cycada} study style transfer from target to source data through deep generative models.
Rather than operating on input data, the feature alignment approach aligns the extracted deep features from both domains
to reduce the feature discrepancy, which can be indicated by maximum mean discrepancy~(MMD)~\cite{rozantsev2018beyond}~\cite{long2017deep}, adversarial neural networks~\cite{tzeng2017adversarial}~\cite{ajakan2014domain}, or reconstruction loss~\cite{ghifary2016deep}. 
However, these traditional UDA approaches require the coexistence of source and target datasets.
This may raise concerns on source-data privacy and can be troublesome when
deploying to resource-constrained devices. 
%because 
% the source datasets are too burdensome to store for resource-constrained devices.
%which may incur data privacy issue and consumes the precious storage of IoT devices in real-world deployments.

% Second cut: source-free that utilizes data compression 
To overcome that, some research works study transforming source data into lightweight forms. 
Works in~\cite{kurmi2021domain}~\cite{hou2021visualizing}~\cite{li2020model} compress the source data into generative models and 
deploy the source model with the data generator to target scenarios for UDA.
However, the deep generator may not protect data privacy~\cite{chakraborty2018adversarial} and still consumes the precious storage of resource-constrained devices.
Other works align target features with the stored statistics of the source feature, such as feature prototype~\cite{qiu2021source}, feature histogram~\cite{ICML2022Measurement}, and batch normalization parameters~\cite{yang2022source}.
Even though, 
they only work for small domain gaps with a trade-off in adaptation quality as compared to the traditional source-based UDA,
because the proposed feature statistics inevitably suffer information loss from the source datasets.

% Third cut: source-free without storage on the source
Source-free UDA further reduces the storage requirement by
adapting the source model with only a set of unlabeled target data,
which is more privacy-preserving and applicable for resource-constrained devices.
Existing studies focus on either input data or feature alignment.
%Existing studies measure the similarity of the target feature to the source. 
The input data-based approaches~\cite{yu2022source}~\cite{xiong2021source}~\cite{karim2023c} learn to
extract domain-invariant features against data augmentation~(e.g., image rotation)
that simulates the domain gap from target to source,
whilst they require target-specific knowledge that is usually unavailable when designing adaptation algorithms.
The feature-based approach studies to measure the similarity of the target feature to the source. 
Works in~\cite{kundu2020universal}~\cite{yang2021generalized} use the information entropy of classification score as an indicator of feature similarity, 
% run target data on source model and optimize the information entropy of their classification scores
where low information entropy indicates source-like features.
% because source features usually have low information entropy of the score.
Other works~\cite{yang2021exploiting}~\cite{zhang2022divide} optimize the compactness of the target features 
because 
the source features are usually clustered or correlated by classification categories.
Nevertheless,
the current source-free approaches either rely on target-specific information or the properties of classification,
% and thus, are hard to be applied or extended to regression tasks. 
which cannot be extended to regression tasks whose target domains are unknown.
In comparison, \sysname{} explores the label distribution of target scenarios to calibrate source models,  
regardless of any classification properties or information of target domains.

%% file: section/method.tex
\section{\sysname{} Design}\label{sec: method}

In this section, we discuss the technical design of \sysname{}.
First, we overview \sysname{} in Section~\ref{subsec: overview} and introduce its confidence classifier in Section~\ref{subsec: confidence_class}.
Then, we discuss 
the label distribution estimator in Section~\ref{subsec: estimator} 
and pseudo-label generator in Section~\ref{subsec: loss}. 

\input{section/overview.tex}

\input{section/estimator.tex}
\input{section/loss.tex}

%% file: section/overview.tex
\subsection{Overview}\label{subsec: overview}

% Problem definition
In the problem,
we have a source regression model~$f_{\theta_s}$
and a bunch of target data~$(x_t,y_t)\in D_t(\subset \mathcal{D})$. 
The parameters of the source model~$\theta_s\subset\Theta$ are learned from a source dataset~$(x_s,y_s)\in D_s(\subset\mathcal{D})$.
The ground truth~(or label) of the target data~$y_t$ exists but is unknown.
Even though the source model~$f_{\theta_s}$ is performing the same task in both source and target scenarios, i.e.,  
$Pr(x|y_s)=Pr(x|y_t)$,
the statistical distribution of inputs (i.e. domain) of both datasets can be different, i.e., $Pr(x_s)\neq Pr(x_t)$, which is
termed domain gap. 
Our objective is to adapt the parameters of the source model $\theta_s$ to be $\theta_t\subset\Theta$,
so the target model $f_{\theta_t}$ minimizes the prediction error on target domain
\begin{equation}
\min \sum_{(x_t,y_t)\in D_t}
\left\lVert f_{\theta_t}(x_t)-y_t\right\rVert_n.
\label{eq: prob_def}
\end{equation}

% Previous works analysis
In the problem setting of source-free UDA, the labels of the target data are unavailable.
We need an alternative objective that complies with Equation~\ref{eq: prob_def}.
%Since the domain gap usually leads to an accuracy drop,
For this, 
previous works try to align accuracy on the target to the source,
where they either simulate or measure from classifiers the domain gap and extract domain-invariant features against it.
However,
they either only work for specific target domains or are designed for classification tasks. 
In this paper, 
we aim to design an adaptation approach for regression models and consider a more practical setting where target domains are unknown in advance,
which calls for a new objective. 
% lacking target-specific information and classification properties, 
% the domain gap is tough to simulate or estimate in the target-agnostic regression task,
% calling for a new approach in place of the previous ones.

% Our intuition
Instead of extracting domain-invariant features,
we directly aim for Equation~\ref{eq: prob_def} by replacing the target label~$y_t$ with a pseudo-label $\hat{y}_t$:
\begin{equation}
\min \sum_{(x_t,y_t)\in D_t}
\left\lVert f_{\theta_t}(x_t)-\hat{y}_t\right\rVert_n,
\label{eq: intuition}
\end{equation}
where the pseudo-label $\hat{y}_t$ is supposed to be closer to the ground truth $y_t$ 
compared with the source model prediction.
% that is, $\left\lVert \hat{y}_t-y_t \right\rVert_n < \left\lVert f_{\theta_s}(x_t)-y_t \right\rVert_n$.
In other words, 
such an adaptation directly works on the label space of target scenarios,
rather than operating on the input or feature space of source models as previous works do.
Even though, it is not intuitive to get the pseudo-label~$\hat{y}_t$. 

% Intuition to calculate pseudo-label
In most machine-learning paradigms, 
the labels that provide supervised information are generally regarded as independent,
while their underlying meanings are essentially correlated in the real world. 
Take, an image-based recognition task, as an example, 
the categories of ``dog'' and ``cat'' naturally resemble each other in front of the label ``boat'',  
though such a pattern is not presented by their one-hot labels. 
Conceptually, the correlation among label classes is referred to as `dark knowledge'' in the field of knowledge distillation~\cite{gou2021knowledge, hinton2014dark, park2019relational}.
By exploring such correlation among label classes, 
works in model compression have successfully equipped small models with the accuracy of large models, especially for deep classifiers~\cite{sun2019patient, polino2018model}. 
Enlightened by knowledge distillation,
we extend the idea of dark knowledge to source-free UDA for regression tasks, 
regarding regression as the classification task with infinite categories of labels. 
Nevertheless, the challenging issues remain: 1) how are the labels of target scenarios correlated, and 2) how to leverage the correlation to get the pseudo-label? 

\begin{figure}[!t]
    \centering
    \begin{minipage}[t]{.23\textwidth}
		\centering
		\includegraphics[width=0.96\textwidth]{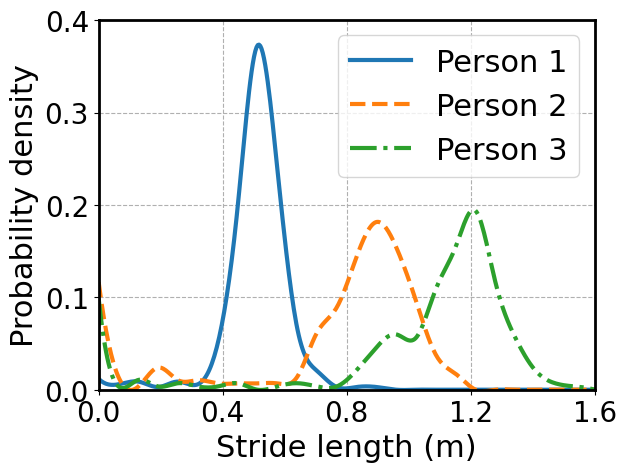}
		\caption{Stride length distribution of different users: label distribution often characterizes target scenarios. }
		\label{fig: label_dist_demo}
	\end{minipage}
\hspace{0.01in}
 \begin{minipage}[t]{.23\textwidth}
		\centering
		\includegraphics[width=0.94\textwidth]{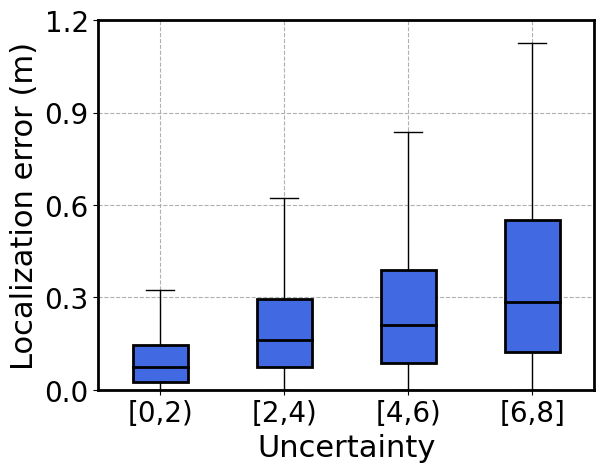}
		\caption{Example of pedestrian dead reckoning: 
                        larger uncertainty tends to indicate larger errors. }
		\label{fig: gauss_dist_demo}
	\end{minipage}
 
\end{figure}

% Explanation of label distribution
We observe that, in many tasks,
target labels are inherently correlated due to the target scenarios.
Specifically, due to the same target scenarios, 
the generation processes of the labels usually share commonalities, such as the same person, site, device, and so on.
Therefore,
just like the input data that all conform to the target domain~(say, cartoon or realistic images), 
the target labels from the same scenario usually form a label distribution that characterizes the scenario. 
To illustrate this,
we show one example in Figure~\ref{fig: label_dist_demo} using the task of stride length estimation,
where the label distribution reflects the walking pattern of the person. 

% Use label distribution to get pseudo-label
From another perspective,
the label distribution of a scenario can be viewed as the prior knowledge of predicting single labels,
which is especially useful when a prediction is uncertain. 
For example, 
if an elder’s stride length mostly falls into a range~(say, 0.5 to 0.8m), 
his/her next stride length is highly likely to be within the range. 
Intuitively, when making a random guess, 
a stride length within the range is expected to be more accurate than the out-of-range one. 
Therefore, 
we can leverage the label distribution to generate pseudo-labels for those uncertain predictions,
which are expected to be more accurate than the original ones. 
In this paper,
we capture the density information of label distribution, 
which serves as the prior knowledge to calculate pseudo-labels. 
More details will be discussed in the following,
where we will cover 
which data need pseudo-labels,
how to estimate the label distribution of target scenarios, 
and how to leverage the distribution to generate the pseudo-labels. 

% because the labels from the same scenario tend to appear in regions that frequently appeared.

\subsection{Confidence Classifier}\label{subsec: confidence_class}

% Why do we need confidence classification? -- 1. to get label distribution; 2. to reduce wrong pseudo label
\indent In this section, we discuss two important considerations to get our intuition down to earth, 
which leads to the design of the confidence classifier.

\emph{How to estimate the label distribution of target scenarios without target labels?}
If the predictions from the source model are accurate,
% these predictions should be close to their ground truths.
we can use these predictions 
to estimate the label distribution of the target.
Unfortunately, 
due to the domain gap between the target and source,
we cannot guarantee all source model predictions on target data are accurate.
Thus, we need a recognition module to differentiate those accurate predictions from all source model predictions on the target data.

The accuracy of predictions is related to the prediction confidence -- 
since the source model usually produces accurate predictions on familiar input data that lead to high prediction confidence,
source model usually shows high confidence in its predictions with high accuracy.
We thus use prediction confidence from source models to recognize the accurate predictions.
% Explain model confidence
Note that confidence~(or uncertainty) estimation for deep learning has been well-studied.
For example, the prediction variance caused by the Dropout layer~\cite{krizhevsky2017imagenet} can be interpreted as prediction confidence~\cite{gal2016uncertainty}.
More uncertainty estimation methods can be found in~\cite{gawlikowski2021survey}.
Since most of the uncertainty estimation approaches are orthogonal to both tasks and model performance,
employing these approaches does not influence the generality of our approach.

\emph{Which kind of target data needs pseudo-labels? }
We consider the label distribution as the scenario's prior knowledge
%We consider exploring the label distribution as an adjustment instead of a prediction
which is independent of the individual pieces of the target data~(or inputs).
% In other words, all target data from the same scenario share the same label distribution.
Thus, 
it fits for calibrating the source model predictions
% when the information from the input data is underutilized. 
% In other words,
when the predictions are uncertain. 
Specifically,
the source model shows low prediction confidence when it has trouble 
analyzing a target input,
indicating a failure to utilize the information from input data.
In this case, 
we use the label distribution 
to calibrate the source model predictions when the prediction confidence is low. 
%those source predictions with low prediction confidence
%should be adjusted by label density.

% How to build the classifier
Considering the two factors, 
we build a confidence classifier to differentiate the target data 
into confident data and uncertain data based on the source model predictions. %, shown as the classification confidence stage in Figure~\ref{fig: sys_diagram}. 
The predictions on the confident data are utilized to estimate the label distribution of the target,
which calibrates the predictions that the source model makes on the uncertain data.
% Deployment
The criterion to 
differentiate the target data is actually related to 
how well the source model learns from the source data.
In other words,
the model's performance on the source data determines its level of confidence in making predictions.
Therefore, 
we differentiate uncertain and confident data based on a threshold of prediction uncertainty~$\tau$ whose value is determined by the model performance on source data.

Specifically,
if a source model learns well from the source dataset,
it should be confident about most of the predictions.
Therefore,
we regard it as a confident prediction if~$\eta$~(proportion) of the source data 
show uncertainty lower than~$\tau$.
% Due to the familiarity, the confident prediction can also be regarded as an accurate prediction if its uncertainty is lower than~$\tau$.
This threshold also applies to target data only if using the same source model. 
It can be determined after the source-model training.  
Finally, we present the confidence classifier by pseudo-code in Algorithm~\ref{alg: conf_classifier}.

\begin{algorithm}[tb]
    \caption{Pseudo code of confidence classifier}
    \label{alg: conf_classifier}
    \textbf{Input}: Target dataset $x_t\in D_t$, source model $f_{\theta_s}$\\
    \textbf{Parameter}: Uncertainty threshold $\tau$\\
    \textbf{Output}: Confident and uncertain data set $SET_C$,~$SET_U$
    \begin{algorithmic}[1]
        \State Initialize~$SET_C$,~$SET_U$
        \For{$x_t$ in $D_t$}
            \State Calculate prediction uncertainty~$u_t$ using~$f_{\theta_s}$
            \If{$u_t>\tau$}
                \State Save~$\left(f_{\theta_s}(x_t), u_t\right)$ to~$SET_C$
            \Else 
                \State Save~$\left(f_{\theta_s}(x_t), u_t\right)$ to~$SET_U$
            \EndIf
        \EndFor
    \newline
    \Return $SET_C$, $SET_U$
    \end{algorithmic}
\end{algorithm}

%% file: section/estimator.tex
\subsection{Label Distribution Estimator}\label{subsec: estimator}

In this section, 
we introduce our design on the label distribution estimator,
which 
delivers a label density map~$M$ of the target scenario
using the source model predictions~$\tilde{y}_t=f_{\theta_s}(x_t)$ 
with prediction uncertainty~$u_t$ from the confident data.
For a concise expression,
we focus on the single-dimensional label and leave its 
extension to the multi-dimensional label at the end of Section~\ref{sec: method}.

% What is the label density map?
We refer to label density as 
the number of labels that appear in a unit region.
Thus, we build a grid~(or discrete) representation of the label density, named label density map.
Formally,
we denote the label density map as a set of label densities:
\begin{equation}
d_i=M(i), 
\label{eq: label_density_map}
\end{equation}
where $i\in \mathcal{N}$ denotes the index for label density $d_i$.
If target label $y_t$ is available, the label density for index~$i$ is
\begin{equation}
d_i=1/D\sum_{k=1}^{K} \mathbbm{1}\left( \frac{y_t^{(k)}-y_0}{g}\in \left[i, i+1\right) \right), 
\label{eq: label_density_map_label_learn}
\end{equation}
where~$\mathbbm{1}(\cdot)$ denotes indicator function, $y_0$ is the smallest label value considered, 
~$g$ is the grid size of the label density map,
~$K$ is the number of the confident data, and
$1/D$ is a normalization term.

% The gap between label and prediction
Unfortunately,  the label of the target data is unavailable,
which requires estimating the label density map.
Based on confident data,
we propose a label distribution estimator.
Specifically, the estimator leverages the correlation between the prediction error and uncertainty
-- the errors tend to be larger with higher uncertainty. 
% because the uncertainty is caused by the unfamiliarity of the source model on the target data,
% which leads to prediction inaccuracy.
This is a natural pattern of deep models~\cite{gal2016uncertainty}, 
of which we further show one supporting example~(from location sensing) in Figure~\ref{fig: gauss_dist_demo}.

\begin{figure}[!t]
\centering
    \includegraphics[width=2.0in]{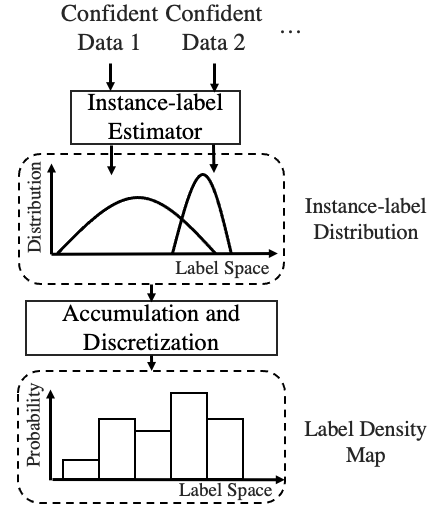}
		\caption{Illustration on label distribution estimator. It first estimates the label distribution of each piece of confident data and then accumulates the estimated instance-label distributions into a label density map.
                        }
		\label{fig: label_estor}
\end{figure}

% Model the relationship
As illustrated in Figure~\ref{fig: label_estor},
the estimator first estimates the label distribution of each piece of confident data.
Then, it accumulates the estimated label distributions as label density.
In detail, for each prediction~$\tilde{y}_t^{(k)}$,
% we estimate its label distribution based on the error-uncertainty correlation.
% Specifically, 
we model the error to be Gaussian distribution so that 
the label conforms to
\begin{equation}
y_t^{(k)}\sim \mathcal{N}\left(y|\tilde{y}_t^{(k)}, \sigma_t^{(k)}\right).
\label{eq: gauss_error}
\end{equation}
This is the instance-label distribution where $\tilde{y}_t^{(k)}=f_{\theta_s}\left(x_t^{(k)} \right)$.
Note that Gaussian distribution is widely adopted in the field of uncertainty estimation~\cite{gawlikowski2021survey, oala2020interval, krizhevsky2017imagenet}.
We use Gaussian distribution because of its popularity and computational efficiency~\cite{zhang2010gaussian}. 
% , especially since \sysname{} often needs to be deployed on edge devices.
% Even though, \sysname{} is orthogonal to the use of error models (more information is in~\cite{gal2016uncertainty,gawlikowski2021survey}). 

To reflect that the error tends to be larger with higher uncertainty, 
the standard deviation should be related to model uncertainty~$u_t$.
Thus, we model their relationship by a function
\begin{equation}
\sigma_t^{(k)}=Q_s\left(u_t^{(k)}\right).
\label{eq: build_q}
\end{equation}
% Model the Q_s
As the source model correlates the prediction and uncertainty,
$Q_s$ can be modeled based on the source dataset before delivering to the target scenario.

We regard the modeling of~$Q_s$ as a curve-fitting problem.
In particular, the standard deviation of the error~$\sigma_t$  entails that around~$68\%$ data whose errors 
should be less than~$\sigma_t$.
Thus, we learn $Q_s$ so that, for each value of uncertainty~$u_t$,
around~$68\%$ predictions in the source datasets have errors lower than~$Q_s(u_t)$.
Nevertheless, $\sigma_t$ is hard to directly determine since~$u_t$ is a continuous variable.
To tackle this,
we divide source data into~$q$ segments according to their prediction uncertainty (similar to Figure~\ref{fig: gauss_dist_demo})
and fit a parameterized curve to those segments 
\begin{equation}
\min \sum_{q'\in q}
\left\lVert Q_s\left(u_s^{(q')}\right)-e_\sigma^{(q')}\right\rVert_n,
\label{eq: q_learn}
\end{equation}
where~$u_s^{(q')}$ is the mean uncertainty of the segment~$q'$, and~$e_\sigma^{(q')}$ is the estimated standard deviation of errors in the segment.
For simplicity, we use a first-order linear regression model
\begin{equation}
        Q_s(u_t)=
         a_0+a_1u_t,
    \label{eq: linear_q_s}
    \end{equation}
where~$a_0$ and~$a_1$ are optimized by least square method~\cite{bjorck1990least}
\begin{equation}
    \begin{cases}
        a_1=\frac{\sum_{q'\in q} u_s^{(q')}e_\delta^{(q')}-|q|\bar{u}_s\bar{e}_\delta}{\sum_{q'\in q} \left(u_s^{(q')} \right)^2-|q|\bar{u}_s^2}, \\
        a_0=\bar{e}_\delta-a_1\bar{u}_s.
    \end{cases}
\end{equation}
Overall, we name it instance-label estimator, which uses each piece of confident data to estimate the instance-label distribution.

% Construction on the label density map
With the function~$Q_s$, 
we are able to model the label distribution of each piece of confident data by Equation~\ref{eq: gauss_error}.
This enables us to assign the label to the density map by probability, i.e.~accumulation and discretization.
In particular, the probability of the~$k^{th}$ label in~$M(i)$ is
\begin{equation}
d_i^{(k)}=\int_{i}^{i+1} S_k\left(y_0+gI\right) dI,
\label{eq: label_prob_est}
\end{equation}
where the Gaussian probability density 
\begin{equation}
S_k(y)=\frac{1}{\sigma_t^{(k)}\sqrt{2\pi}}\exp{\left( -\frac{\left(y-\tilde{y}_t^{(k)}\right)^2}{2\left(\sigma_t^{(k)}\right)^2} \right)},
\label{eq: prob_density}
\end{equation}
and the standard deviation $\sigma_t^{(n)}=Q_s\left(u_t^{(k)}\right)$.
Totally, the label density map can be estimated by
\begin{equation}
d_i=1/D\sum_{k=1}^K d_i^{(k)},
\label{eq: label_density_map_label_learn_overall}
\end{equation}
% d_i=1/D\sum_{k=1}^K \mathbbm{1}\left( \frac{y_t^{(k)}-y_0}{g}\in \left[i, i+1\right) \right)d_i^{(k)},
based on Equation~\ref{eq: label_density_map_label_learn} with probability calculated from Equation~\ref{eq: label_prob_est}.
%Based on the function~$Q_s$, in the end,
Finally, we present the pseudo-code of label distribution estimator in Algorithm~\ref{alg: estimator}.

\begin{algorithm}[tb]
    \caption{Pseudo code of label distribution estimator}
    \label{alg: estimator}
    \textbf{Input}: Confident data set~$SET_C$\\
    \textbf{Parameter}: Grid size~$g$, label value range~$y\in\left[y_0, y_m\right]$, function~$Q_s$ from Equation~\ref{eq: build_q}\\
    \textbf{Output}: Label density map~$M$
    \begin{algorithmic}[1] %[1] enables line numbers
        \State Calculate the number of grids~$J=\left\lfloor \frac{y_m-y_0}{g}\right\rfloor$
        \State Initialize $M(j)$ for $j=1,2..J$ 
        \For{$\left(f_{\theta_s}(x_t), u_t \right)\in  SET_C$}
            \State Calculate $\sigma_t=Q_s(u_t)$
            \For{$j=1,2..J$}
                \State $S(y)=\frac{1}{\sigma_t\sqrt{2\pi}}\exp{\left( -\frac{\left(y-f_{\theta_s}(x_t)\right)^2}{2\left(\sigma_t\right)^2} \right)}$
                \State $M(j)=M(j)+\int_{j}^{j+1} S\left(y_0+gI\right) dI$ 
            \EndFor
        \EndFor
        \For{$j=1,2,...J$}  \Comment{Normalization}
            \State $M(j)=M(j)/\left|SET_C\right|$
        \EndFor
    \newline
    \Return $M$
    \end{algorithmic}
\end{algorithm}

\iffalse

\begin{figure}[!t]
\centering
    \includegraphics[width=1.8in]{figure/gauss_dist.png}
		\caption{Example of the relationship between model uncertainty and error based on pedestrian dead reckoning: 
                    localization error tends to grow with larger localization uncertainty.}
		\label{fig: gauss_dist_demo}
\end{figure}

$d_i=\sum_{n\in N} \mathbbm{1}\left( y_t^{(n)}\in \left[y_0+gi, y_0+g(i+1)\right) \right), $

\FOR{$(\hat{y}_s, u_t)$ in BUFFER1}
            \STATE Name $\hat{y}_s$ as $\hat{y}_t$
            \IF{$u_t<\tau$}
                \STATE Save $\hat{y}_t$ to BUFFER2
            \ELSE
                \STATE Initialize $\tilde{y}_t$ as a zero vector
                \FOR{$(\hat{y}_s, u_t)$ in BUFFER1}
                    \IF{$\left\lVert \hat{y}_s-\hat{y}_t\right\rVert_n<r $}
                        \STATE $\tilde{y}_t=\tilde{y}_t+\hat{y}_s/u_t$
                    \ENDIF
                \ENDFOR
                \STATE Normalize $\tilde{y}_t$ by $u_t$ and save it to BUFFER2
            \ENDIF
        \ENDFOR
        \STATE Train $f_s$ using $D_t$ with pseudo labels from BUFFER2 and output $f_t$

\fi

%% file: section/loss.tex
\subsection{Pseudo-label Generator}\label{subsec: loss}
% Introduction
In this section, 
we first introduce how to create the pseudo-label using the label density map $M$.
Then, we discuss evaluating the credibility of each pseudo-label.
Both parts contribute to the loss function that 
supervises the adaptation training for the source model.

\begin{figure}[!t]
\centering
    \includegraphics[width=2.5in]{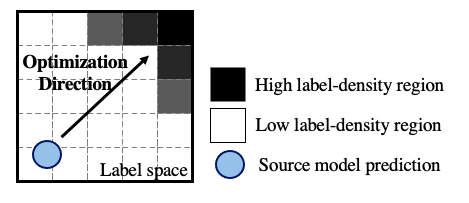}
		\caption{Illustration of the basic idea of pseudo-label generator.}
                % source model predictions should be close to the high-density regions on the label density map. }
		\label{fig: generator}
\end{figure}

% Pseudo label
\emph{How to generate pseudo-labels using label density map?}
The basic idea of the pseudo-label generator is illustrated in Figure~\ref{fig: generator}.
We stick to the grid representation of label density map and denote the grid range as
\begin{equation}
Y_i=y_0+g[i, i+1).
\label{eq: label_range}
\end{equation}
As mentioned, 
we regard the label distribution of confident data as the prior knowledge of the label of uncertain data. 
For an uncertainty data $x_t^{(j)}$, we estimate the posterior label distribution as the joint distribution of prior knowledge and its instance-label distribution
\begin{equation}
Pr\left(y_t^{(j)}\in Y_i\right)=Pr\left( y\in Y_i|\tilde{y}_t^{(j)}, u_t^{(j)} \right)\times Pr(Y_i),
\label{eq: pseudo_label_prob}
\end{equation}
where~$\tilde{y}_t^{(j)}=f_{\theta_s}\left(x_t^{(j)}\right)$.
On the right-hand side, 
we % use the~$Q_s$ from Equation~\ref{eq: build_q} to
model the first probability as a Gaussian distribution similar to Equation~\ref{eq: gauss_error} and the second probability using
label density map $Pr(Y_i)=M(i)$. 
% In total, they contribute to the left-hand side, the individual label distribution of a piece of target data,
% whose grid represents the probability to include the label.
%that considers the adjustment of the scenario's prior knowledge, i.e., the label density map, on the predictions of the uncertain data.

Based on the posterior label distribution,
we generate the pseudo-label that should be close to the grids with high probability.
Instead of selecting the grid with the highest probability,
we calculate the pseudo-label by interpolating grids according to their probability:
\begin{equation}
\hat{y}_t^{(j)}=1/Y\sum_i Pr\left(y_t^{(j)}\in Y_i\right)\bar{Y_i},
\label{eq: pseudo_label_cal}
\end{equation}
where~$\bar{Y_i}$ denotes the center of the grid, and~$1/Y$ is the normalization term. 
Through such an interpolation,
the generated pseudo-label is naturally close to the dense grids when the label density map shows a clear local trend.
Otherwise, it will be close to the source model prediction.
This avoids causing accuracy degradation when the prior knowledge is not informative. 

To understand the relationship between the estimated label distribution of confident data~$Pr(y_{con})$ and uncertain data~$Pr(y_{unc})$,
we present~$Pr(y_{unc})$ by the joint distribution of N independent samplings
\begin{equation}
    Pr(y_{unc})=\prod_{i=1}^N Pr\left(y_{unc}^{(i)} \right).
\end{equation}
By considering Equation~12, their relationship can be represented as
\begin{equation}
    \log Pr(y_{unc})=N\log Pr(y_{con})+\sum_{i=1}^{N} \log Pr(y_{unc}^{(i)}).
\end{equation}
Intuitively, $Pr(y_{con})$ serves as the prior knowledge for estimating~$Pr(y_{unc})$, while they do not have to be the same. 

% Weighting
\emph{Should we equally trust all pseudo-labels?}
We regard the label density map as a preference~(or prior knowledge of the scenario)
when the source model is uncertain about its predictions. % the information from the target data is not sufficiently utilized.
Thus, we should trust more about the pseudo-label when the source model prediction is not confident, and vice versa. 
In order to evaluate the credibility of the pseudo-labels, 
we normalize the confidence of the source model predictions.
% Since the source model is familiar with the source data, 
% we quantify the information utilized from the inputs as prediction uncertainty regarding the uncertainty threshold 
In particular,
we use the confidence threshold~$\tau$ as a reference to normalize the confidence of source model prediction 
\begin{equation}
I_d=\frac{\tau}{u_t^{(j)}}.
\label{eq: weight_info}
\end{equation}
%where $u_t^{(j)}>\tau$ because the uncertainty lower than the threshold would be considered 
%as a sufficiently-utilized case, i.e., prediction on the confident data.

% We should also take into account the credibility of the pseudo-labels themselves --
Also, we assign a higher credibility to pseudo-labels when a clear trend is formed in the label density, 
% the resulting pseudo-labels calculated from it have higher credibility.
% we should trust more on the pseudo-labels 
% when the label density map shows a strong trend in the prediction's locality.
On one hand,
a clear trend means that the local label densities of a prediction~(in label space) should not be evenly distributed.
This can be achieved by the interpolation method in Equation~\ref{eq: pseudo_label_cal} 
because the uniform distribution of the local density will render the pseudo-label close to the prediction.
% predictions are likely to be adjusted to a local maximum in the label density map. 
On the other hand, the location where the pseudo-labels are calculated should have a high label density.
We denote such a feature by local mean density~$\bar{d}_l$ regarding the global mean density~$\bar{d}_i$
\begin{equation}
I_l=\frac{\bar{d}_l}{\bar{d}_i}.
\label{eq: weight_label}
\end{equation}
Here, we regard the locality as the grids whose centers are within three standard deviations from the prediction, i.e.
\begin{equation}
\left\lVert \bar{Y}_i-\tilde{y}_t^{(j)}\right\rVert_n<3\sigma_t^{(j)},
\label{eq: locality_deviation}
\end{equation}
% Loss function formula
and the credibility of the pseudo-label is
\begin{equation}
\beta_t=\frac{I_l}{I_d}.
\label{eq: weight_total}
\end{equation}
We use it as the weight of the pseudo-label in the adaptation training. 

% Adaptation training
Overall, 
the loss function for the adaptation training is
% This leads to the proposed target label loss function
\begin{equation}
\mathcal{L}_{ada}= \sum_{\substack{x_t\in D_t, \\ u_t>\tau}}
\beta_t  \mathcal{L}\left(f_{\theta_s}(x_t), \hat{y}_t\right),
\label{eq: loss_func}
\end{equation}
where the pseudo-label is calculated from Equation~\ref{eq: pseudo_label_cal},
the loss weight~$\beta_t$ is from Equation~\ref{eq: weight_total}, and~$\mathcal{L}$ is task-dependent. %can be any task-related loss function that reduces the discrepancy between labels and model predictions.
%Based on the loss function in Equation~\ref{eq: loss_func}, adaptation training is the typical training process of deep models.
% Use confident data to train as well
Besides training on the uncertain data,
we suggest as well involving the confident data in the adaptation training using the pseudo-label~$\hat{y}_t=\tilde{y}_t$.
Because the confident data also belongs to the target data, involving them in the training data facilitates the model adapting to the target scenario and avoids the catastrophic forgetting issue~\cite{de2021continual}
where deep models may forget previous knowledge when learning new ones.
The pseudo-code of the pseudo-label generator is presented in Algorithm~\ref{alg: generator}.

% Multi-dimensional cases
Finally, 
we discuss extending the approach to tasks with \emph{multi-dimensional labels}.
It mainly distinguishes from the case of single-dimensional labels by requiring a label density map
with a multi-dimensional index~$i\in \mathcal{N}^m$ where $m$ is the label dimension. 
This leads to a multivariate Gaussian distribution in Equation~\ref{eq: gauss_error},
which requires estimating the covariance matrix for Equation~\ref{eq: q_learn}.
For simplicity, 
we suggest treating label dimensions as independent
if they are not coupled by the loss function during the training process.

\begin{algorithm}[tb]
    \caption{Pseudo code of pseudo-label generator}
    \label{alg: generator}
    \textbf{Input}: Label density map~$M$, uncertain data set~$SET_U$, grid size~$g$, minimum label value~$y_0$, grid number~$J$\\
    \textbf{Parameter}: Uncertainty threshold $\tau$, function~$Q_s$\\
    \textbf{Output}: Pseudo-label set~$SET_P$
    \begin{algorithmic}[1] %[1] enables line numbers
        \State Initialize~$SET_P$
        \State $\bar{d}_i=\sum_{j=1}^JM(j)/J$ \Comment{Calculate global mean density}
        \For{$\left(f_{\theta_s}(x_t), u_t \right)\in SET_C$}
            \For{j=1,2,...,J}
                \State $VAR_W=0$, $VAR_Y=0$, $\beta_t=0$
                \State Initialize $SET_M$
                \State $\sigma_t=Q_s(u_t)$
                \State $y_m=y_0+(j+0.5)g$ \Comment{Calculate grid center}
                \If{$\left|y_m-f_{\theta_s}(x_t) \right|<3\sigma_t$}
                    \State $S(y)=\frac{1}{\sigma_t\sqrt{2\pi}}\exp{\left( -\frac{\left(y-f_{\theta_s}(x_t)\right)^2}{2\left(\sigma_t\right)^2} \right)}$
                    \State $VAR_W+=M(j)\times \int_{j}^{j+1} S\left(y_0+gI\right) dI$
                    \State $VAR_Y=VAR_Y+y_m\times VAR_W$
                    \State Save~$M(j)$ to $SET_M$
                \EndIf
            \EndFor
            \State $\hat{y}_t=VAR_Y/VAR_W$ \Comment{Calculate pseudo-label}
            \State $\beta_t=\frac{\bar{d}_i\times u_t}{\tau}\times \frac{\sum_{m\in SET_M}m}{\left|SET_M \right|}$ \Comment{Calculate credibility}
            \State Save~$(\hat{y}_t, \beta_t)$ to~$SET_P$
        \EndFor
        \newline
        \Return $SET_P$
    \end{algorithmic}
\end{algorithm}

%% file: section/exp.tex
\section{Illustrative Experimental Results}\label{sec: exp}

In this section,
we demonstrate illustrative experimental results to verify \sysname{},
where we introduce the experimental setting in Section~\ref{subsec: exp_set}
and present illustrative results in Section~\ref{subsec: result}.

\subsection{Experimental Setting}\label{subsec: exp_set}
% Dataset
To verify \sysname{},
we first experiment it with two regression tasks  --
pedestrian dead reckoning~\cite{yan2019ronin}  and image-based people counting~\cite{zhang2016single} --
because their target domains are usually heterogeneously different,
and the applications often require a source-free adaptation due to 
storage and privacy concerns.
Then, we additionally verify \sysname{} on two prediction tasks -- 
California housing price prediction~\cite{calihouseprice} and New York City taxi trip duration prediction~\cite{nyc-taxi-trip-duration}. 
The two tasks further validate \sysname{}'s generality to different tasks. 

\emph{Pedestrian dead reckoning}~(PDR)~\cite{yan2019ronin} 
is a task of location sensing.
It aims to estimate the user's walking trajectory 
using the phone-mounted IMU sensors, 
specifically, the accelerometer and gyroscope.
We employ RoNIN~\cite{yan2019ronin} as a baseline model to adapt,
which is a state-of-the-art PDR model based on temporal-convolutional neural network~(TCN).
The model focuses on 2D trajectories,
where we model the two dimensions independently.
In the experiment,
we adapt the baseline model to 25 users separately,
wherein 15 users have contributed to the source datasets but perform differently in the tests~(small domain gap), 
and the other 10 users are completely unseen by the baseline model~(large domain gap).

Different users have different walking behaviors with random carriage states of the phone, 
causing heterogeneous domain gaps.
Each user may contribute one or multiple trajectories --
user in the seen group provide, on average, 250m trajectories,
and that of the unseen group is 500m.
To verify that \sysname{} is applicable to not only the target data that have been adapted but all data from the target scenario, for each user,
we use $80\%$ trajectories for adaptation and the rest for testing.
Note that the labels are unavailable in both adaptation and testing.

The evaluation for PDR focuses on how well the model recovers the trajectory.
In this paper, we evaluate the PDR model by two metrics:
\begin{itemize}
    \item {\emph{Step error (STE)}}: 
    The model outputs a displacement vector using IMU signals every two seconds~(one step).
    We measure the Euclidean distance between model output and ground truth in each step 
    and average them over a trajectory by
    \begin{equation}
        STE=
        1/J\sum_{j\in J} \left\lVert y_j - \tilde{y}_j \right\rVert_2,
    \label{eq: step_error}
    \end{equation}
    where the trajectory has $J$ steps;
    \item {\emph{Relative trajectory error (RTE)}}~\cite{yan2019ronin}: 
    RTE measures the localization error in terms of trajectory
    \begin{equation}
        RTE=
         \left\lVert \sum_{j\in J} y_j - \sum_{j\in J} \tilde{y}_j \right\rVert_2
    \label{eq: rte}
    \end{equation}
    with an aligned starting point between the estimated trajectory and the ground-truth trajectory.

\end{itemize}

% To show that \sysname{} is general to different tasks, 
We also experiment \sysname{} on \emph{image-based people counting}~\cite{wang2020nwpu} which
counts the number of people from single images.
We use MCNN~\cite{zhang2016single} as our baseline model, which is 
a classic and well-recognized people-counting approach based on convolutional neural network.
In our experiment, the baseline model is trained on Part-A (482 images) of the Shanghaitech dataset~\cite{zhang2016single} and are adapted to Part-B (716 images) of it,
where the two parts are differentiated by scenes and people densities. 
The resolution of each image is~$768\times 1024$. 
% In the experiment,
% we adapt the model to the Part-B dataset.
Similar to the PDR experiment, we use $80\%$ data for adaptation and the rest for testing.
We follow the original paper to evaluate the experimental results by mean squared error~(MSE) 
and mean absolute error~(MAE). 

To verify the generality of \sysname{} to different tasks,
we additionally apply it to predict California housing price~\cite{calihouseprice} and
New York City taxi trip duration~\cite{nyc-taxi-trip-duration}. 
Generally, the two tasks are using the provided features~(such as house age and pickup date) to predict house prices in California and taxi trip duration in New York. 
To form domain gaps,
we separate the two datasets spatially since both the house price and taxi trip duration are related to (house or take-off) location. 
Specifically,
we separate California as coastal~(target) and non-coastal~(source) areas according to~\cite{woloszyn2022advancing} 
and New York as Manhattan~(target) and non-Manhattan~(source) areas. 
We employ a MLP-based model~\cite{poongodi2022new} as baseline and
evaluate the two tasks by  
mean squared error~(MSE) and rooted mean squared logarithmic error~(RMSLE), as provided by their datasets. 

% Comparison
We compare \sysname{} with the following state-of-the-art schemes:
\begin{itemize}
    \item {\emph{MMD-based UDA~(MMD):}} Work in~\cite{long2017deep} proposes a traditional source-based UDA approach using source data. 
    It measures the domain gap by MMD and aligns them in the feature space;
    \item {\emph{ADV-based UDA~(ADV):}} Work in~\cite{tzeng2017adversarial} proposes a traditional source-based UDA approach using source data.
    It leverages a pre-trained adversarial neural network to bridge the domain gap in feature space;
    \item {\emph{UDA without source data~(Datafree):}} Work in~\cite{ICML2022Measurement} conducts UDA without using source data.
    Instead, it
    stores source feature distribution via a soft histogram and 
    regards the feature distribution as a domain gap.
    \item {\emph{Augmentation-based source-free UDA (AUGfree):}} Work in~\cite{xiong2021source} is a source-free UDA approach based on data alignment.
    It requires a known domain gap and simulates the gap by data augmentation, where the domain-invariant features are extracted. 
    In the experiment, we follow the original paper and employ the variance perturbation as the augmentation method.
\end{itemize}

% Magic numbers
In the experiment,
we use the Dropout mechanism to calculate model uncertainty.
Uncertainty is presented by the standard deviation of predictions from twenty samplings with a dropout rate of 0.2.
% For the confidence classifier in Section~\ref{subsec: confidence_class}, we empirically set $\eta=0.9$.
To reduce randomness, 
we repeat each experiment five times and report the average result. 
% each verification is repeated by five times where the averaged result is reported.
Unless particularly specified,
we show the results on the adaptation set. 
% of the uncertain dataset for a fair evaluation.

\input{section/result.tex}

%% file: section/result.tex
\begin{figure}[!t]
\centering
    \includegraphics[width=2.6in]{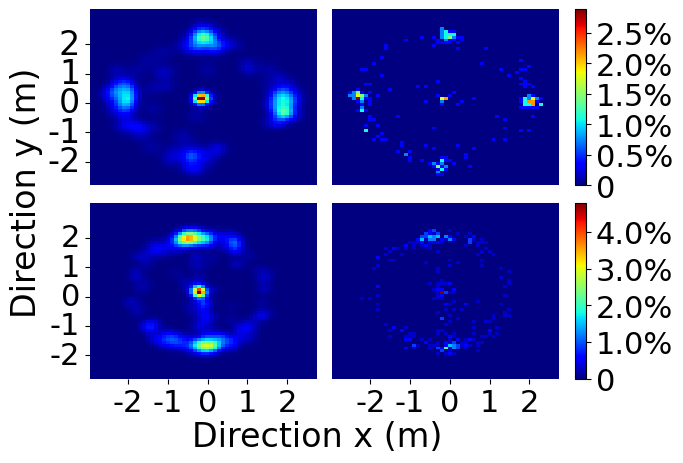}
		\caption{Visualization of the estimated~(left) and true~(right) label density map based on two PDR users. }
		\label{fig: vis_density_map}
\end{figure}

\begin{figure*}[tp]
    \centering
	\begin{minipage}[t]{.23\textwidth}
		\centering
		\includegraphics[width=\textwidth]{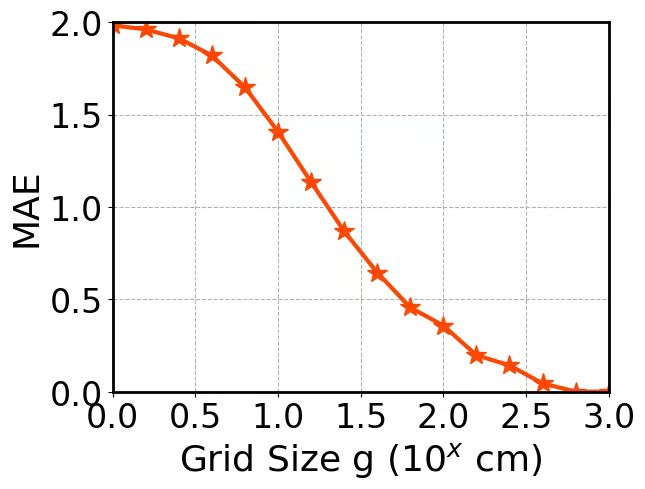}
		\caption{Error of label distribution estimator varies with grid size:
                    a larger grid size leads to a lower estimation error. }
		\label{fig: map_mae}
	\end{minipage}
	\hspace{0.1in}
    \begin{minipage}[t]{.23\textwidth}
		\centering
		\includegraphics[width=\textwidth]{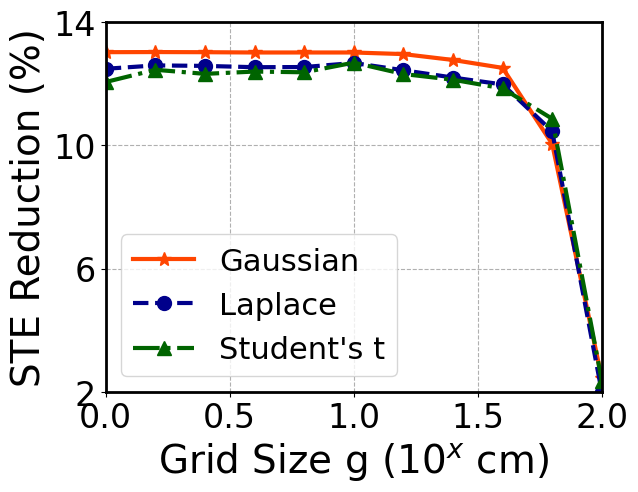}
		\caption{Pseudo-label error varies with grid size: a large grid size is not preferred.}
		\label{fig: grid_accu}
	\end{minipage}
 	\hspace{0.1in}
    \begin{minipage}[t]{.23\textwidth}
		\centering
		\includegraphics[width=\textwidth]{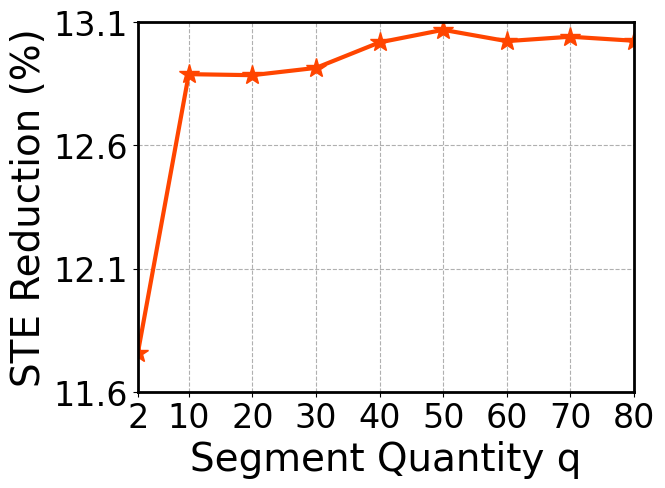}
		\caption{Pseudo-label error varies with segment quantity~$q$: a too small~$q$ is not preferred.}
		\label{fig: accu_q}
	\end{minipage}
 	\hspace{0.1in}
    \begin{minipage}[t]{.23\textwidth}
		\centering
		\includegraphics[width=\textwidth]{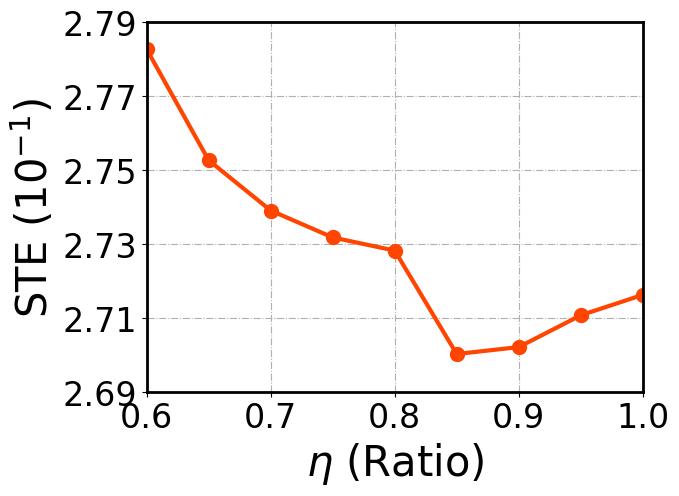}
		\caption{Pseudo-label error varies with the ratio~$\eta$. }
		\label{fig: eta_accu}
	\end{minipage}
\end{figure*}

\subsection{Illustrative Results}\label{subsec: result}

In this part,
we experiment on the system parameters using PDR in Section~\ref{subsubsec: sys}
and show performance of \sysname{} with the comparison schemes in Section~\ref{subsubsec: pdr}.
Then, we extend the experiments and analysis to people counting in Section~\ref{subsubsec: counting}.
Evaluations of the two prediction tasks are in Section~\ref{subsubsec: prediction}.
A failure case is analyzed in Section~\ref{subsubsec: failure}.

% Verification of system parameters
\subsubsection{Study on System Parameters}\label{subsubsec: sys} 
We study how \sysname{} performance varies with its system parameters. 
Unless specified, the experiments are on the seen group of PDR
using the identical grid size on the two label dimensions of PDR label.

% Visualization
We first visualize the estimated label density maps 
and compare with their ground truth, using two sample users from PDR.
In Figure~\ref{fig: vis_density_map}, 
% Description of features from ground truths
from the ground truths of the label density maps,
both label density maps display ring-shaped patterns in the high-density grids,
which indicates the users' regular walking speeds.
Also, the clusters on rings indicate the users' walking patterns. 
From the figure, 
the estimated label density maps accurately capture the ring-shaped pattern and clustering information of the high-density grids.
The larger ring of the upper figure shows that the walking speed of the user tends to be larger than the other user.
And, the clustered regions of the high-density grids indicate that 
the upper user is more likely to make sharp turns than the other one.
This confirms the effectiveness of the label distribution estimator and 
justifies the use of the estimated label density map to calibrate source models.

% MAE on the label density map
% The label distribution estimator generates the label density map using Equation~\ref{eq: label_density_map_label_learn_overall}.
To verify the label distribution estimator~(Equation~\ref{eq: label_density_map_label_learn_overall}), 
we present the mean absolute error (MAE) of the estimated label density map in Figure~\ref{fig: map_mae}.
As shown in the figure, the MAE converges to MAE=2/0 with extremely small/large grid sizes.
This is because larger grids ease the estimation task, and vice versa.
For instance,
an extremely large grid would involve all target data in both the estimated and ground truth maps,
leading to the same label density.
Nevertheless,
we will provide an explanation as to why using a large grid is not recommended in the subsequent analyses.

% Grid size g against accuracy
In Figure~\ref{fig: grid_accu},
we show how pseudo-label accuracy varies with grid size based on different distribution forms of error models. 
First, 
there is no significant difference among different error models,
which verifies that \sysname{} is compatible with different distribution forms, 
as long as it shows larger errors for high prediction uncertainty. 
In terms of pseudo-label accuracy,
the figure suggests a small grid size, 
while it may lead to low accuracy as in Figure~\ref{fig: map_mae}.
This is because the grid interpolation of \sysname{} (in Equation~\ref{eq: pseudo_label_cal}) makes it robust to the estimation error, while 
the performance will only degrade with an extremely large grid. 
% an extremely large grid may influence the performance due to its coarse grid resolution.
Overall,   
the system performance is not sensitive to the choice of the grid size.
Even though, 
an extremely small grid size is not preferred. 
Specifically,
the computation complexity of constructing a label density map is $O(n/g)$ based on $n$ pieces of confident data with grid size $g$,
indicating that it consumes more computing resources to construct a label density map with smaller grid size. 
Since the accuracy flattens off when the grid size reduces, 
there is no need for a small grid size. 

% Segmentation q against accuracy
In figure~\ref{fig: accu_q}, we set the grid size to be $10cm$ and investigate how pseudo-label accuracy varies with segment quantity $q$ in Equation~\ref{eq: q_learn}. 
% This equation models the relationship between model uncertainty and prediction error. 
The figure shows that the pseudo-label accuracy quickly converges with a small~$q$.
Therefore, only a few segments can capture the relationship between model uncertainty and prediction error. 
% indicating that the system's performance is not highly sensitive to the choice of segment quantity. 
The convergence also shows that \sysname{} works with a wide range of~$q$.
% This is because $Q_s$ from Equation~\ref{eq: build_q} 
% inherently estimates a parameter of a distribution, which is less reliant on modeling accuracy. 
We empirically set $q=40$ for the following experiments.

\begin{figure*}[tp]
    \centering
    \begin{minipage}[t]{.21\textwidth}
		\centering
		\includegraphics[width=\textwidth]{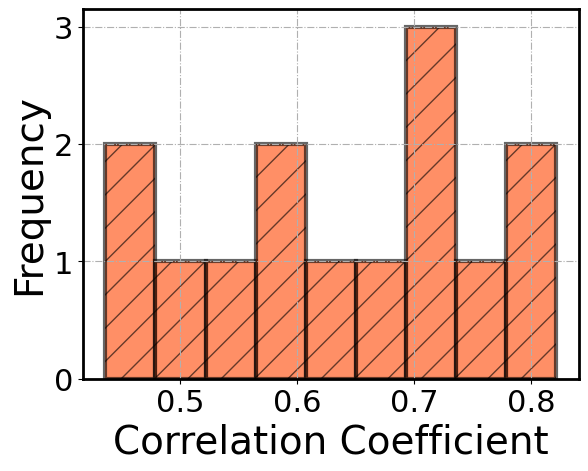}
		\caption{Distribution of the correlation coefficient between credibility~$\beta_t$ and prediction error over different users.}
		\label{fig: beta_correlation}
	\end{minipage}
   	\hspace{0.1in}
 \begin{minipage}[t]{.26\textwidth}
		\centering
		\includegraphics[width=\textwidth]{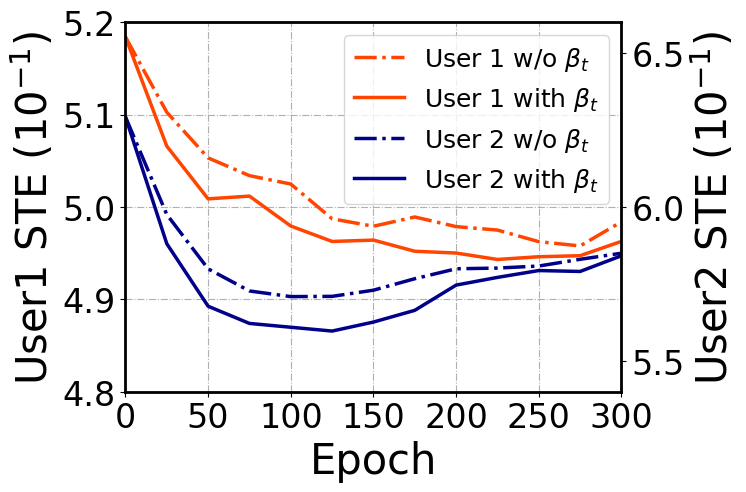}
		\caption{Ablation study on the credibility~$\beta_t$. }
		\label{fig: ablation_study}
	\end{minipage}
   	\hspace{0.1in}
 \begin{minipage}[t]{.23\textwidth}
		\centering
		\includegraphics[width=\textwidth]{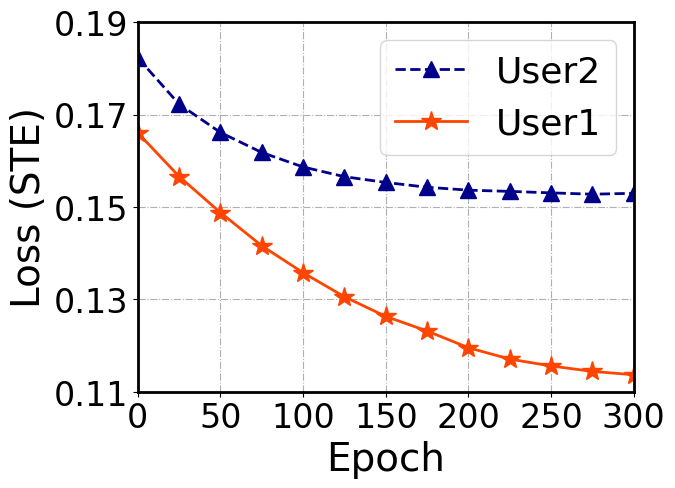}
		\caption{Learning curves of adaptation training: 
                    early stop when the rate of error reduction slows down. }
		\label{fig: user_learn_curve}
	\end{minipage}
 	\hspace{0.1in}
    \begin{minipage}[t]{.21\textwidth}
		\centering
		\includegraphics[width=\textwidth]{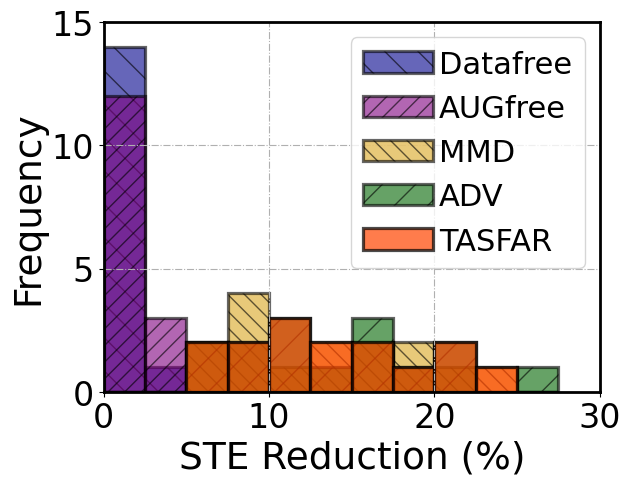}
		\caption{Comparison on STE reduction rate. } %: \sysname{} outperforms the existing source-free UDAs and performs comparably with source-based UDAs  on various users.}
		\label{fig: seen_ada_hist}
	\end{minipage}
\end{figure*}

% \eta against accuracy
In Figure~\ref{fig: eta_accu}, 
we study how to select the confidence ratio~$\eta$ for the confidence classifier as discussed in Section~\ref{subsec: confidence_class}. 
% Describe the curve
The figure shows how pseudo-label error varies with~$\eta$,
where the pseudo-label error decreases when~$\eta$ is less than 0.9.
% Explain the curve
As explained in Section~\ref{subsec: confidence_class},
a small~$\eta$ leads to a small confidence threshold~$\tau$ so that
the accurate predictions may be considered as the uncertain ones. 
Also,
a too large~$\eta$ may decrease the numbers of uncertain data such that no data are available for adaptation. 
% How to select eta
Even though, 
the figure shows a wide range of~$\eta$ to use. 
In this paper, 
we set~$\eta$ to be $0.9$. 
% we usually select~$\eta$ from 0.8 to 1.0 so that
% the confident data can be used to improve the prediction accuracy on the uncertain data.

% beta verification
We validate the pseudo-label credibility~$\beta_t$ (from Equation~\ref{eq: loss_func}) in Figure~\ref{fig: beta_correlation}.
For trajectory data~(with multiple steps) of each person, 
we calculate the Pearson correlation coefficient of~$\beta_t$ and the pseudo-label accuracy and
summarize them as a probability distribution function~(PDF) over different users. 
As shown in the figure, 
the coefficients of all users exhibit a positive correlation, 
where most users' correlations are larger than 0.5. 
Therefore,
\sysname{} will assign large weights to the accurate pseudo-labels~(in Equation~\ref{eq: loss_func}), which avoid generating low-quality pseudo-labels that cause accuracy degradation. 
% Such a high correlation confirms the use of the confidentiality of the pseudo-label $\beta_t$ as the loss weights .

% Ablation study
We further conduct an ablation study of $\beta_t$ in Figure~\ref{fig: ablation_study}.
With or without using the weight~$\beta_t$, the figure compares the STE varies with epochs in the adaptation training.
Both curves show lower STEs with $\beta_t$, 
while the gaps are reduced with more training epochs.
This is because the pseudo-labels with larger~$\beta_t$ tend to be more accurate than those with small weights.
The model would stress more on the pseudo-labels with large~$\beta_t$ in the beginning because of the large weights.
This explains the gap shown in the two curves. 
As the number of epochs increases,  
the gaps are reduced because the training losses of these pseudo-labels~(with large~$\beta_t$) are reduced when
the model starts to focus on the ones with less accurate pseudo-labels. 
% This is because the confident pseudo-labels are more likely to be accurate than the less confident ones,
Thus, we should employ an early stop to improve adaptation performance. 

% How to enforce early stop
As the adaptation training process is automatic and unsupervised, we study the early stop issue in Figure~\ref{fig: user_learn_curve}
We show the learning curve of the same users as in Figure~\ref{fig: ablation_study}.
In the figure, 
both curves show the regular patterns of deep model training: 
the speed of the training loss drops gradually reduces as the epoch increases. 
% This is because the training processes also involve certain data of users, as explained in Section~\ref{subsec: loss}.
The significant training loss drops, at small epochs,
shows that the adaptation training 
is bridging the gaps between source model predictions 
and pseudo-labels with large weights~$\beta_t$~(from Equation~\ref{eq: loss_func}).
Therefore, 
the change in the loss-dropping speed indicates a changing focus from the large~$\beta_t$ to the smaller one.
So, we can early stop the adaptation training when the loss-dropping speed is significantly reduced, 
i.e.~epoch~250 of user~1 and~epoch~100 of user~2. 
This also agrees with the satisfactory stopping epochs from Figure~\ref{fig: ablation_study}.
% As a large epoch may focus the training more on the pseudo-labels with small weights,  
% we early stop the training process when the training loss drop is significantly reduced,  
% which also avoids the overfitting issue. 

\subsubsection{Performance Analysis in PDR}\label{subsubsec: pdr}

In this part,
we analyze the experimental results on PDR and compare \sysname{}
with the comparison schemes.
Unless particularly specified, we demonstrate results on adaptation data.

\begin{figure*}[tp]
    \centering
	\begin{minipage}[t]{.23\textwidth}
		\centering
		\includegraphics[width=\textwidth]{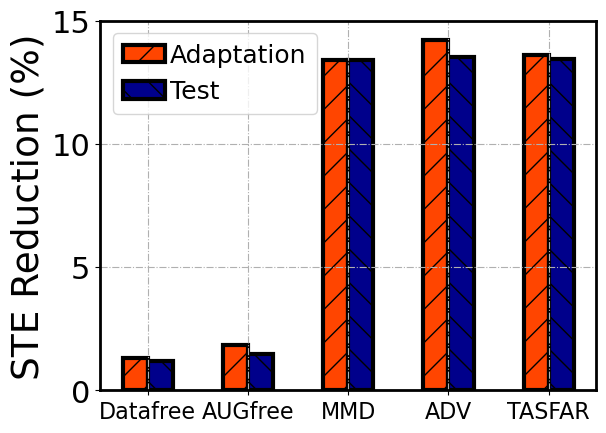}
		\caption{Comparison on STE reduction between adaptation and test sets.} % : the result on the test set is consistent to that of adaptation set.}
		\label{fig: unseen_bar}
	\end{minipage}
   	\hspace{0.1in}
 \begin{minipage}[t]{.24\textwidth}
		\centering
		\includegraphics[width=\textwidth]{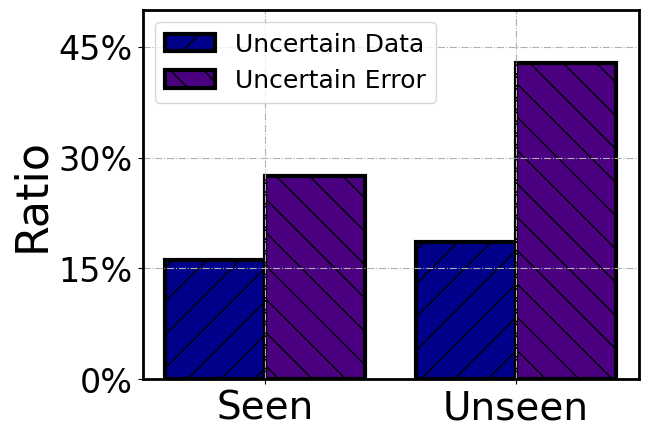}
		\caption{The ratio of uncertain data and errors in the seen and unseen group. }
                    % The small proportion of uncertain data is responsible for a significant portion of the prediction errors observed in the target data. }
		\label{fig: uncertain_prop}
	\end{minipage}
     \hspace{0.1in}
    \begin{minipage}[t]{.22\textwidth}
		\centering
		\includegraphics[width=\textwidth]{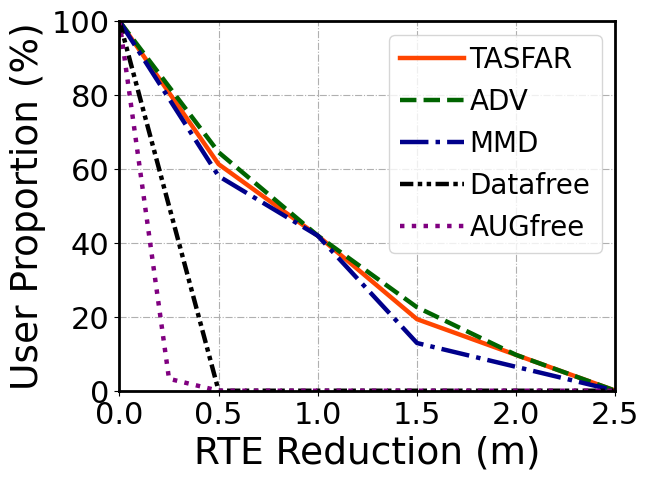}
		\caption{ How many users' RTE from the seen group are reduced? }
            % \sysname{} shows a comparable RTE reduction rate to the source-based UDA approaches. }
		\label{fig: seen_cdf}
	\end{minipage}
  	\hspace{0.1in}
    \begin{minipage}[t]{.22\textwidth}
		\centering
		\includegraphics[width=\textwidth]{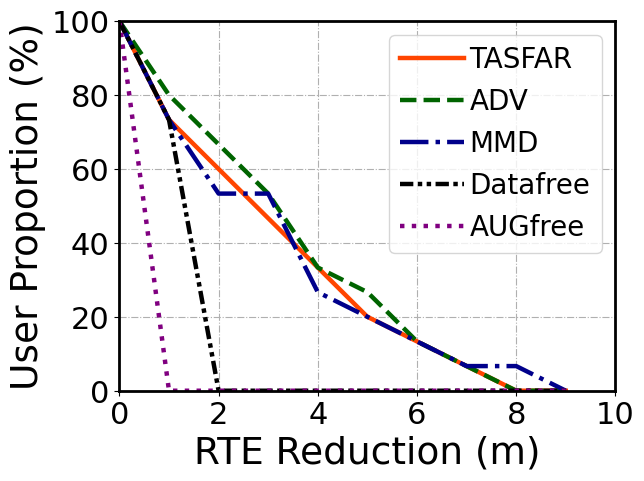}
		\caption{How many users' RTE from the unseen group are reduced? }
            % \sysname{} performs comparably with source-based UDA on both small and large domain gaps.}
		\label{fig: unseen_cdf}
	\end{minipage}
\end{figure*}

% Seen Adaptation STE: target-agnostic
We first evaluate the STE reduction in the seen group 
and show the reduction distribution over the individual user in Figure~\ref{fig: seen_ada_hist}.
The figure shows \sysname{} achieves similar error reduction compared with source-based UDA approaches, i.e., MMD and ADV,
while the improvement from other source-free approaches is insignificant.
Datafree can only achieve small improvements because it merely aligns domains in terms of feature statistics.
The adaptation performance of AUGfree varies across different users because its augmentation only fits a few users.
In comparison,
the STE of each person is significantly reduced by applying \sysname{} because it
directly calibrates the source model using label distribution of the target scenarios.
Considering that different users have different signal distributions,
this experiment has verified that \sysname{} is practical and general to heterogeneous target domains.

% Adaptation to test STE
We verify the performance consistency in adaptation and test sets in Figure~\ref{fig: unseen_bar}.
\sysname{} achieves an averaged STE reduction of $13.6\%$ in the adaptation set and $13.4\%$ in the test set,
% the averaged STE reduction results are compared between adaptation and test sets.
and all schemes show similar error reductions between the two sets. 
Firstly, 
the schemes are not accessible to labels of both adaptation and test sets. 
% Therefore, the available information shows no difference to test data.
Secondly
data from both sets are generated from the same domain,
where the test data distribution is similar to that of the adaptation.
This explains the consistent performance of \sysname{} in both adaptation and test sets and validates
that the adaptation can be achieved by using a group of data from target domains.

% From uncertain data to whole data
Figure~\ref{fig: uncertain_prop} shows the ratio of uncertain data and their prediction errors regarding the whole dataset.
Due to the domain gaps between the target and source,
the uncertain data ratios of both seen and unseen groups are larger than~$\eta=0.9$.
The ratio of the unseen group~$18.6\%$ is larger than that of the seen group~$16.2\%$ due to its larger domain gap.
From the figure,
the error ratios are much larger than the data ratios in both groups
because the errors are mainly incurred by uncertain data.
Therefore, \sysname{} only pseudo-labels the uncertain data, though,
it can achieve commendable adaptation performance because the uncertain data group is the main source of the inaccuracy.
% Additionally,
% training on the accurate pseudo-labels of the uncertain data 
% facilitates the target model to extract better features from the target domain,
% which not only adapts the target model to uncertain data but also potentially improves its accuracy on the confident data. 

% seen test RTE
Besides using STE to show model accuracy on the uncertain data,
% While STE is directly showing the accuracy of model outputs,
we show the RTE of both confident and uncertain data in the test set.
Specifically, we show how many localization errors are reduced over the test trajectories in Figure~\ref{fig: seen_cdf}. 
The figure shows the numbers~(in ratio) of trajectories whose error reduction is more than a threshold~(x-axis),
where \sysname{} achieves~$0.92m$~(or 7\%) average error reduction for trajectories with an average length of $50m$.
This is comparable with the source-based UDA,
outperforming other source/data-free approaches.
Also, the result conforms to the conclusion drawn from the STE experiments.
Note that the localization error of PDR is temporally dependent, 
where the location of the next step depends on the last one.
So, the errors sometimes cancel each other's bias over the trajectories. 
Therefore, it is possible that Datafree can outperform AUGfree in RTE while performing worse in STE. 
% As the error accumulation, the error reduction could be increasingly significant with longer trajectories,
% which will be shown in the unseen test sets.

% Unseen Test RTE
We study the performance of \sysname{} to both small~(seen) and large~(unseen) domain gaps in Figure~\ref{fig: unseen_cdf}.
As users from the unseen group are not exposed to the source model in model training,
the domain gaps of the unseen group are larger than those of the seen group.
As the errors of PDR are cumulative,
the error reduction is more significant in longer trajectories. 
From the figure,
\sysname{} still shows comparable RTE reduction with the source-based UDA approaches.
It reduces around $3.13m$ of RTE for trajectories with an average length of $100m$.
This validates that \sysname{} is capable of handling both small and large domain gaps in terms of the input distribution
because it explores label space that is decoupled from the input space.

\subsubsection{Performance Analysis in People Counting}\label{subsubsec: counting}
\begin{figure}
    \centering
    \includegraphics[width=0.4\textwidth]{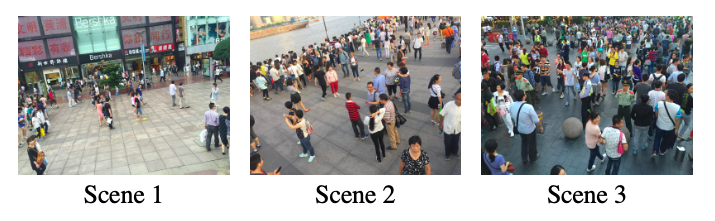}
    \caption{Sample images of sites from the people-counting dataset~\cite{zhang2016single}.}
    \label{fig: scene}
\end{figure}

To show that \sysname{} can work with multiple signal forms, 
we conduct experiments of image-based people counting. 
To capture the properties of target scenarios,
we apply \sysname{} to images belonging to the same sites~(streets) separately from the test dataset.
The sample images from the three sites are shown in Figure~\ref{fig: scene},
where scene~3 tends to be more crowded than the others from our observation.

\begin{figure*}[tp]
    \centering
    \begin{minipage}[t]{.22\textwidth}
		\centering
		\includegraphics[width=\textwidth]{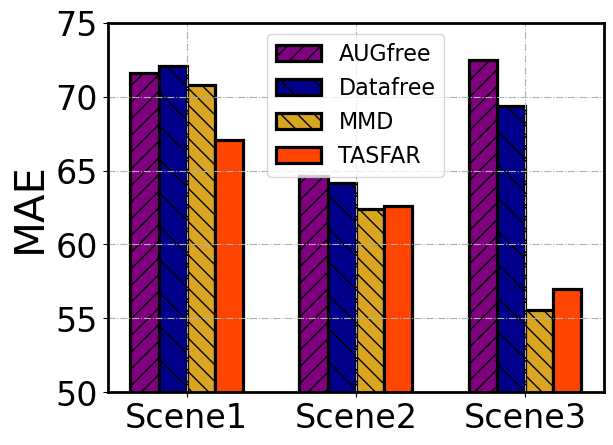}
		\caption{Comparison of the different scenes of people counting on the test set.  }
		\label{fig: counting_scene}
	\end{minipage}
 \hspace{0.1in}
 \begin{minipage}[t]{.23\textwidth}
		\centering
		\includegraphics[width=\textwidth]{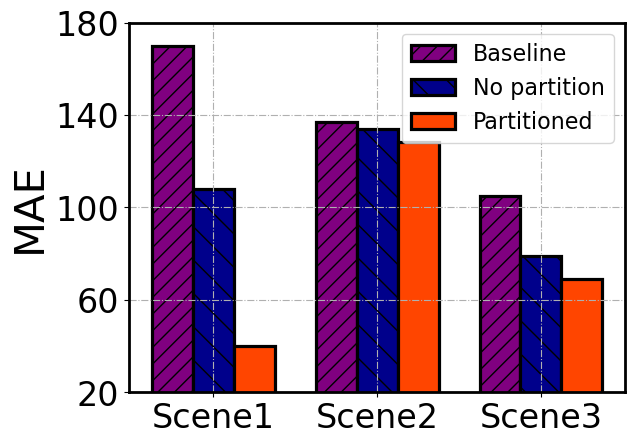}
		\caption{\sysname{}'s performance with or without partitioning the test data.  }
		\label{fig: partition}
	\end{minipage}
 \hspace{0.1in}
 \begin{minipage}[t]{.27\textwidth}
		\centering
		\includegraphics[width=\textwidth]{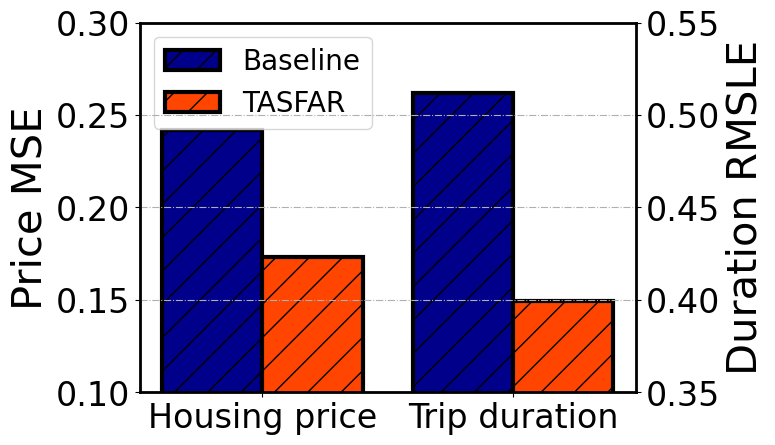}
		\caption{\sysname{}'s performance on the two prediction tasks.  }
		\label{fig: exp_predict}
	\end{minipage}
 \hspace{0.1in}
 \begin{minipage}[t]{.20\textwidth}
		\centering
		\includegraphics[width=\textwidth]{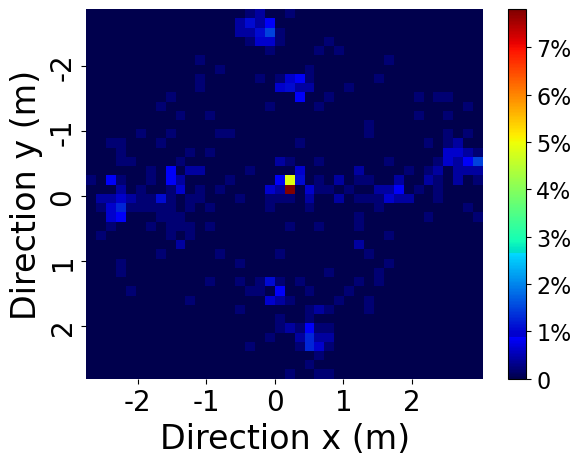}
		\caption{Label distribution with two users in PDR.  }
		\label{fig: exp_fail}
	\end{minipage}
 
\end{figure*}

% Table
\begin{table*}
    \centering
        \caption{Comparison on crowd counting. \sysname{} performs comparably with the traditional source-based UDA approaches on the adaptation set, uncertain data from the adaptation set, and test set.}
    \begin{threeparttable}[b]
    \begin{tabular}{ccccccccccccccc} %{lllllllllll}
        \hline
        ~  & \multicolumn{2}{c}{Adaptation (whole)}  & \multicolumn{2}{c}{\bf Error Reduction (\%)}& \multicolumn{2}{c}{Adaptation (uncertain)}  & \multicolumn{2}{c}{\bf Error Reduction (\%)}   & \multicolumn{2}{c}{Test} & \multicolumn{2}{c}{\bf Error Reduction (\%)}\\
         Scheme &    MAE      &   MSE  &    MAE      &   MSE  &    MAE      &   MSE    &   MAE & MSE    &   MAE & MSE   &   MAE & MSE   \\
        \hline
         Baseline & 56.4 & 86.8 & - & - & 114.4 & 143.8 & - & - & 71.7 & 141.5 & - & -\\ 
         MMD\tnote{*} & 51.5 & 82.2 & 8.7 & 5.3 & 80.1 &  96.7  &    29.9 & 32.7 & 59.6 & 110.9 & 16.8 & 21.5 \\
         ADV\tnote{*} & 52.1 & 81.9 & 7.6 & 5.6 &  76.4 &   97.2  &  33.2  &  32.4 & 59.7 & 111.0 & 16.7 & 21.6\\
        \hline
         AUGfree  & 56.3 & 86.9 & 0.2 & 0 & 113.6 & 142.5 & 0.7 & 0.9 & 71.5 & 141.0 & 0.2 & 0.3\\
        Datafree & 56.2 & 86.1 & 0.4 & 0.8 & 108.1 &  134.5   & 5.5 & 6.5  & 69.5 & 134.1 & 3.0 & 5.2 \\
         {\bf \sysname{}} & 52.4 & 80.3 & {\bf 7.0} & {\bf 7.5} & 74.0 & 97.5  & {\bf 35.3} & {\bf 32.2} & 59.9 & 107.4& {\bf 16.5} & {\bf 24.1} \\
        \hline
    \end{tabular}
     \begin{tablenotes}
      \item[*]  Source-based UDA approach
   \end{tablenotes}
    \label{tb: counting}
    \end{threeparttable}
\end{table*}

% Table
In Table~\ref{tb: counting},
we compare the experimental results on the adaptation and test set.
% wherein we use the uncertain set of adaptation data for further demonstration.
% which comes to similar conclusions to the experiments in PDR.
The source model~(baseline) performs worse on the uncertain set than on the whole adaptation set because of the high prediction uncertainty.
Although  all schemes reduce more errors on the uncertain set than the whole adaptation set due to the large error,
\sysname{} significantly outperforms the other source-free approaches in both MSE and MAE
and achieves comparable results with the source-based UDA approaches.
% The test and adaptation data are coming from the same target domain.
% The baseline model performs worse on the test data, though,
% the results after adaptation on the test set, in terms of error reduction rate, 
After adaptation, 
the experimental results of the test set also 
come to a consistent conclusion:
Datafree only reduces errors slightly and
AUGfree does not perform well on people counting because its augmentation approach misfits the task,
while \sysname{} achieves~$16.5\%$ and~$24.1\%$ error reduction in terms of MAE and MSE that are comparable with the source-based UDA approaches. 

% Comparison of the three scenes
We further compare \sysname{} with the other approaches on different scenes in Figure~\ref{fig: counting_scene}.
We only show MMD because it performs similarly to ADV.
Similarly to the results in Table~\ref{tb: counting},
\sysname{} achieves comparable performance with the source-based UDA on all three scenes, 
outperforming AUGfree and Datafree. 
Interestingly,
\sysname{} outperforms the existing source-free approaches in scene~2 and~3, and 
surpass them by a large margin in scene~1.
This is because the crowded scene~3 maintains a stable pedestrian stream, 
forming a prominent feature in label distribution.
In all,
the accuracy improvement in all three scenes has
verified that \sysname{} can work with different crowd scenes. 

% Comparison with and w/o partition
In Figure~\ref{fig: partition}, we discuss \sysname{}'s performance 
without partitioning the test dataset by scene. 
% In the figure,
% we compare \sysname{}'s performance in terms of MAE with or without partitioning the 
% test dataset by scenes.
% To illustrate the difference, 
% we use the uncertain data of the adaptation set in the experiment.
From the figure,
\sysname{} shows better performance in all three scenes when their adaptation sets are partitioned.
This is because data from the same scenes are correlated by the target scenarios, 
providing prominent features in the label distribution that are leveraged by \sysname{}. 
On the contrary, 
fusing data with multiple scenes may corrupt the features of each target, 
degrading the adaptation performance of \sysname{}.
Even though, 
\sysname{} can still achieve good performance without partitioning because the crowd density of the Part B dataset is inherently correlated. 
% Therefore,
% it is possible to improve the performance of \sysname{} by partitioning the test data,
% which is discussed in Section~\ref{sec: discuss}.

\subsubsection{Performance Analysis in Prediction Tasks}\label{subsubsec: prediction}
To verify the generality of \sysname{} in different tasks, 
we further show in Figure~\ref{fig: exp_predict} its performance on two prediction tasks.
On the target regions, 
\sysname{} has reduced 22\% MSE and 28\% RMSLE separately in predicting housing price and trip duration. 
Since (house and take-off) location is a key factor of housing price and trip duration, 
the baseline models that are learned from one district cannot perform well in another district. 
Even though, the housing prices and trip duration in the target district are naturally correlated. 
\sysname{} captures such correlation and improves the accuracy for the target district. 

\subsubsection{Failure Case Analysis}\label{subsubsec: failure}
Finally,
we show a failure case in the PDR task where the target model, calibrated by \sysname{}, is only marginally better than the source model. 
Specifically, 
we manually balance the target data by using two users' data as the target, upon which \sysname{} only reduces around 1\% STE. 
The performance is similar to those of other source-free approaches. 
To analyze it, we visualize target label distribution in Figure~\ref{fig: exp_fail}. 
As shown, the two users have different step lengths and walking patterns, 
so the label distribution displays a double-ring shape that differentiates the single-person case in Figure~\ref{fig: vis_density_map}. 
However, 
the label distribution of one user usually cannot serve as the prior knowledge of the other,
resulting in a failure of adaptation. 
To avoid causing accuracy degradation,
the \sysname{} would generate pseudo-labels that are close to the source-model predictions~(due to the double-ring shape) and assign small weights to adaptation loss since the label densities are spread out over the map. 
Ideas to tackle such cases are further discussed in Section~\ref{sec: discuss}.

\iffalse

\begin{figure*}[tp]
    \centering
	\begin{minipage}[t]{.30\textwidth}
		\centering
		\includegraphics[width=\textwidth]{figure/exp_g_mae.png}
		\caption{Mean absolute error~(MAE) of the label density map varies with grid size~$g$: 
                    a larger grid size leads to a lower estimation error. }
		\label{fig: map_mae}
	\end{minipage}
	\hspace{0.15in}
    \begin{minipage}[t]{.29\textwidth}
		\centering
		\includegraphics[width=\textwidth]{figure/exp_g_accu.png}
		\caption{Pseudo-label error varies with grid size~$g$: a large grid size is not preferred.}
		\label{fig: grid_accu}
	\end{minipage}
 	\hspace{0.15in}
    \begin{minipage}[t]{.30\textwidth}
		\centering
		\includegraphics[width=\textwidth]{figure/exp_q_ste.png}
		\caption{Pseudo-label error varies with segment quantity~$q$: a small~$q$ is not preferred.}
		\label{fig: accu_q}
	\end{minipage}
\end{figure*}

First,
\sysname{} 
significantly outperforms other source/data-free approaches, in terms of error reduction on MAE and MSE,
and achieves comparable results with the source-based UDA approaches.
Second,
the adaptation and test sets show consistent improvement,
indicating the feasibility to  learn from the adaptation set. 
Last, 
except for the AUGfree whose target-based augmentation misfits the people counting task,
most schemes show larger improvements than their results in the PDR 
because of the different tasks and source models.
Overall, the experiment of people counting further corroborates  \sysname{}'s practicability and generality to different  regression tasks.

\fi

%% file: section/conclude.tex
\section{Conclusion}\label{sec: conclude}
The traditional source-based unsupervised domain adaptation~(UDA) 
uses both unlabeled target data and the training dataset~(on the source domain) to
overcome the domain gap between the target and source.
To protect source data confidentiality and reduce storage requirements,
source-free UDA replaces source data with a source model and adapts it to the target domain.
Previous source-free UDA approaches 
% operate on the input or feature space of the source model and hence
measure and bridge domain gaps in input-data or feature space of the source model, which
only works for specific domain gaps or classification tasks.
% cannot be applied to regression tasks without knowing the target domains beforehand. 

In this paper,
we propose, for the first time, 
a \emph{\textbf{t}}arget-\emph{\textbf{a}}gnostic \emph{\textbf{s}}ource-\emph{\textbf{f}}ree domain \emph{\textbf{a}}daptation approach termed \emph{\textbf{\sysname{}}} for \emph{\textbf{r}}egression tasks.
\sysname{} is based on the observation that the target label, 
like target data that all conform to the target domain,
also originates from the same target scenario. 
Therefore,
in contrast to previous source-free UDA approaches, % that operate on input or feature space of the source model,
\sysname{} directly estimates the label distribution of the target scenario and uses it to calibrate source models. 
% {\LARGE DRAW PARALLEL WITH "IN CONTRAST" directly optimize the source model predictions on target data}. 
% Based on prediction confidence,
% {\LARGE PLEASE DESCRIBE COMPLETELY AND BRIEFLY HOW THE WHOLE SCHEME WORKS HERE \sysname{} first estimates a label density map  to represent the label distribution,
% and then uses the map to pseudo-label the source model predictions.}
Specifically,
\sysname{} classifies the target data into confident and uncertain data
and proposes a label distribution estimator, based on the confident data, to estimate the target label distribution, represented as a label density map.
Then, a pseudo-label generator utilizes the label density map to pseudo-label the uncertain data, 
which is used to fine-tune the source model based on supervised learning. 

To validate \sysname{}, we have conducted extensive experiments on four regression tasks, namely, pedestrian dead reckoning (using the inertial measurement unit), image-based people counting from single images, and two prediction tasks. 
We compare \sysname{} with state-of-the-art source-free UDA and source-based UDA approaches. 
The experimental results show that \sysname{} significantly
outperforms the existing source-free UDA
with around~$14\%$ and~$24\%$ reduction in localization error and mean absolute error (MSE) in the pedestrian dead reckoning of different users and people counting with various crowd scenes, respectively.  
In the two prediction tasks, \sysname{} reduces 26\% of the prediction errors. 
Without the need for any source data, its performance outperforms the previous source-free UDA approaches and
is notably comparable with the source-based UDA approaches.

% {\LARGE THE NUMBER OF REFERENCES IS ON THE LOW SIDE.  MAY ADD MORE TO 4X REFERENCES.}

%% file: section/discuss.tex
\section{Future Work}\label{sec: discuss}
% {\LARGE WHAT IS THE POINT OF THIS SECTION?  IF IT IS ON FUTURE WORKS, PLEASE MORE IT AFTER CONCLUSION "CONCLUSION AND FUTURE WORKS".  IF IT IS ON MODEL ASSUMPTIONS, IT SHOULD BE PUT IN THE INTRODUCTION.}
% How to interpret the properties of scenarios
To achieve source-free domain adaptation for regression tasks on agnostic target domains, 
\sysname{} explores the label properties that originate from target scenarios themselves, 
such as environmental features, behavioral patterns of users, cyclic events of the scenes, and so on. 
This observation makes \sysname{} well-suited for adaptation in real-world scenarios, 
where the label distributions are naturally imbalanced because of the heterogeneous target scenarios. 
Consequently, its performance gain is not so marked in tasks 
where the target data comes from multiple sources or where labels are manually balanced, 
such as those datasets for data competitions. 
\sysname{} may achieve only minimal accuracy improvement 
on such tasks since their scenario properties may be corrupted or intentionally reduced.

% Future works on exploring  scenario properties in specific applications
One direction of future works can focus on how to partition test data so as to better utilize the characteristics of the target scenario.
% in order to achieve better adaptation performance,
This partition may depend on task-specific knowledge.
When applying \sysname{} to a specific task, 
we can partition the target data, according to the task-specific knowledge, into several parts,
in which we pseudo-label the uncertain data independently. 
% conduct \sysname{} independently on these parts.
For example, in a surveillance-based people counting, 
\sysname{} may perform better if we treat the morning and evening as two target scenarios. 
% indicating a satisfactory temporal feature to partition target data in this application.
From this perspective,
\sysname{} may serve as a general framework to incorporate more task-specific knowledge to achieve better adaptation performance on real-world applications. 

% Discussion on classification
Finally, 
we discuss our outlooks of applying \sysname{} to classification tasks, though it is specifically designed for regression models. 
Technically,
\sysname{} may be straightforwardly applied to classification tasks. % because of its weaker assumption compared with the classification-based approaches. 
Without leveraging classification properties, however,
\sysname{} by itself is not expected to show advantages over those approaches in classification tasks. 
Despite so,
it is possible to combine \sysname{} with other classification-based approaches as a plug-in module. 
Specifically,
\sysname{} may be used to explore the correlation among label classes of a classification task
and generate soft pseudo-labels for uncertain data.
Such kind of information~(namely dark knowledge) has been successfully verified in the field of knowledge distillation. 
We thus believe it may be useful in source-free domain adaptation, 
which can be an interesting future work to study.